\documentclass{article}

\usepackage[square,sort,comma,numbers]{natbib}

\usepackage[preprint]{neurips_2025}

\usepackage[utf8]{inputenc} 
\usepackage[T1]{fontenc}    
\usepackage{hyperref}       
\usepackage{url}            
\usepackage{booktabs}       
\usepackage{amsfonts}       
\usepackage{nicefrac}       
\usepackage{microtype}      
\usepackage{xcolor}         

\usepackage{amsmath} 
\usepackage{times}
\usepackage{mathptmx}
\usepackage{tikz}
\usetikzlibrary{shapes.geometric, arrows, positioning}
\usepackage{stfloats}
\usepackage{graphicx}
\usepackage{pifont}  
\usepackage{cleveref} 
\usepackage{subfigure}
\usepackage{multirow}
\usepackage{makecell}
\usepackage{verbatim}
\usepackage{wrapfig} 
\usepackage{caption} 
\usepackage{subcaption}
\usepackage{minitoc}       
\usepackage{algpseudocode}

\usepackage{colortbl}
\definecolor{LightGreen}{rgb}{0.85, 1, 0.85} 
\definecolor{LightRed}{rgb}{1, 0.7, 0.7}

\usepackage{algorithmicx}
\usepackage{algorithm}
\usepackage{algpseudocode}

\usepackage[textsize=tiny,textwidth=0.6in]{todonotes}
\usepackage[textsize=tiny,textwidth=0.6in]{todonotes}
\usepackage{bm}
\newcommand{\allnotes}[1]{}
\renewcommand{\allnotes}[1]{#1}

\title{Hypernetworks for Model-Heterogeneous Personalized Federated Learning}

\author{%
  Chen Zhang$^{1,2}$ \quad Husheng Li$^{1,2}$ \quad Xiang Liu$^{3,*}$ \quad Linshan Jiang$^{3,*}$ \quad Danxin Wang$^{1,2}$\\[2mm]
  $^{1}$\,the Qingdao Institute of Software, College of Computer Science and Technology,\\
  China University of Petroleum (East China), China\\
  $^{2}$\,the Shandong Key Laboratory of Intelligent Oil \& Gas Industrial Software\\
  $^{3}$\,National University of Singapore\\[1mm]
  \begin{tabular}{c}
    \texttt{zhangchen@upc.edu.cn}, \texttt{z23070167@s.upc.edu.cn}, 
    \texttt{wangdx@upc.edu.cn}\\
    \texttt{liuxiang@comp.nus.edu.sg},
    \texttt{linshan@nus.edu.sg}\\
    * indicates the corresponding author
  \end{tabular}
}
\begin{document}

\maketitle

\begin{abstract}
Recent advances in personalized federated learning have focused on addressing client model heterogeneity. However, most existing methods still require external data, rely on model decoupling, or adopt partial learning strategies, which can limit their practicality and scalability. In this paper, we revisit hypernetwork-based methods and leverage their strong generalization capabilities to design a simple yet effective framework for heterogeneous personalized federated learning. Specifically, we propose MH-pFedHN, which leverages a server-side hypernetwork that takes client-specific embedding vectors as input and outputs personalized parameters tailored to each client’s heterogeneous model. To promote knowledge sharing and reduce computation, we introduce a multi-head structure within the hypernetwork, allowing clients with similar model sizes to share heads. Furthermore, we further propose MH-pFedHNGD, which integrates an optional lightweight global model to improve generalization. Our framework does not rely on external datasets and does not require disclosure of client model architectures, thereby offering enhanced privacy and flexibility. Extensive experiments on multiple benchmarks and model settings demonstrate that our approach achieves competitive accuracy, strong generalization, and serves as a robust baseline for future research in model-heterogeneous personalized federated learning.
\end{abstract}

\doparttoc %
\faketableofcontents %

\section{Introduction}
Federated learning (FL) has been widely applied in various fields, such as intelligent transportation~\cite{zhou2024adaptive,wang2022ai}, healthcare~\cite{murmu2024reliable,ouyang2023harmony,nguyen2022federated}, and recommendation systems~\cite{feng2024robust,yuan2023federated,guo2021prefer}. However, a single global model cannot meet all clients' needs due to non-IID data. To address this, personalized federated learning (pFL)~\cite{10.5555/3294996.3295196,t2020personalized,deng2020adaptive} emerges, aiming to craft personalized models for clients while enabling knowledge sharing under cross-device settings~\cite{McMahan2016CommunicationEfficientLO}, thus better matching their specific tasks and data distributions.

In practice, devices participating in pFL are often heterogeneous, as they usually have different computational resources~\cite{chai2020tifl,
shin2024effective}, communication capabilities~\cite{Caldas2018ExpandingTR, 10.5555/3495724.3496904, Shah2021ModelCF}, and model architectures~\cite{li2019fedmd,zhu2021data,wu2024exploring}, which complicates the challenges that pFL faces in scenarios of model heterogeneity~\cite{Chen2023EfficientPF}. To address the limitations of the model heterogeneous pFL (MH-pFL),  several methods have been proposed by researchers, including partial training~\cite{diao2021heterofl,alam2022fedrolex, hong2022efficientsplitmixfederatedlearning,10.1145/3643832.3661880}, federated distillation~\cite{Zhang2023TowardsDK, wang2024dfrd, liu2025adapter,10734591} and model decoupling~\cite{xu2023personalized,pmlr-v139-collins21a,arivazhagan2019federated,liang2020thinklocallyactglobally}.

However, modeling decoupling~\cite{10.1145/3545008.3545073,
10.1145/3581783.3611781,yi2023pfedes} separates local models into feature extractors and classifiers; this low-level knowledge sharing may hinder client collaboration and negatively impact performance. Partial training methods~\cite{NEURIPS2021_6aed000a,Ilhan_2023_CVPR,lee2024recurrentearlyexitsfederated} allow clients to select sub-models for local training. However, differences in client model architectures, data distributions, and resource conditions can lead to misalignment in parameter counts and feature spaces, causing suboptimal performance. Current federated distillation approaches~\cite{Gong_2021_ICCV,itahara2021distillation,Wang_Yan_Wang_Wang_Shu_Cheng_Chen_2025} often depend on a universal or synthetic dataset for knowledge integration. Such additional datasets may limit the applicability of the method and make the performance dependent on the quality of these datasets. Hence, an MH-pFL approach is needed to effectively resolve these issues while ensuring privacy and computational efficiency.

In this work, we revisit hypernetwork~\cite{ha2017hypernetworks}, a model that generates personalized parameters for another neural network, and propose a novel MH-pFL framework, the \textit{\textbf{M}odel-\textbf{H}eterogeneous \textbf{P}ersonalized \textbf{Fed}erated \textbf{H}yper\textbf{N}etwork} (\texttt{MH-pFedHN}), which uses a hypernetwork as a knowledge aggregator to enable knowledge fusion among heterogeneous clients. In our approach, MH-pFedHN customizes embedding vectors for each client based on the number of parameters required by the client's model, whereas clients with a similar number of parameters use the same customized embedding vectors.
In addition to the feature extractor from traditional hypernetwork shared across all clients to support generalizable feature learning, a shared head is created for clients with the same embedding vectors to generate client-specific parameters for heterogeneous models, which allows the server to generate parameters for multiple models with the similar number of parameters in a single pass, increases efficiency and promotes knowledge fusion. This fine-grained mapping mechanism enhances the server's expressive and generalization abilities.

To further learn cross-client knowledge and enhance performance, we propose \textit{\textbf{MH-pFedHN} with \textbf{G}lobal \textbf{D}istillation} (\texttt{MH-pFedHNGD}), with an additional lightweight but effective global model. This plug-in component on the server side is directly generated by our hypernetwork using the global customized embedding vectors.
Compared to MH-pFedHN, MH-pFedHNGD leverages a global model to aggregate updates from clients, introducing one more round lightweight update mechanism for the hypernetwork, enabling it to learn a more comprehensive data distribution and more generalizable features. Meanwhile, the global model can serve as a teacher model~\cite{hinton2015distilling,jeong2018communication} to assist in the training via knowledge distillation on the client side, thereby enhancing the knowledge acquisition ability of client-specific models and balancing personalization with generalization. 

Both MH-pFedHN and MH-pFedHNGD are data-free and preserve the structural privacy of personalized heterogeneous models. They are the first efficient frameworks designed to leverage hypernetworks for solving MH-pFL problems, and hold great promise for future applications. Our main contributions are as follows.

\begin{itemize}  
\item We propose MH-pFedHN, a personalized FL method based on hypernetworks and specifically designed for heterogeneous models. This method allows the server to utilize customized embedding vectors and the shared head tailored to clients' needs to generate parameters without disclosing the model architecture to the server, thereby enhancing privacy protection.

\item We propose MH-pFedHNGD, which integrates a plug-in component lightweight global model. This approach enhances the hypernetwork's learning and generalization capabilities and allows personalized client models to learn from a global model with knowledge distillation efficiently, resulting in significant performance improvements.

\item We evaluate our MH-pFedHN and MH-pFedHNGD for multiple tasks on four popular datasets. 
The experiments demonstrate that our method exceeds state-of-the-art performance against baselines across all tasks, highlighting the effectiveness and generalizability. 
\end{itemize}  

\section{Related Work}

\subsection{Personalized Federated Learning with Hypernetworks}

Hypernetworks are methods that use one network to generate weights for other neural networks~\cite{ha2017hypernetworks}, which are widely used in various machine learning applications~\cite{von2020continual,suarez2017language,nirkin2021hyperseg, beck2024recurrent, ruiz2024hyperdreambooth}. In pFL, hypernetworks are used to generate personalized model parameters from clients' embedding vectors~\cite{shamsian2021personalized,zhu2023layer,scott2024pefll} and output the weight ratio during aggregation~\cite{ma2022layer}. This method has shown effectiveness in systems with diverse data and few-shot learning scenarios~\cite{sendera2023hypershot}. 

However, these methods are mainly designed to alleviate the problem of statistical heterogeneity, so that clients can obtain personalized models. The work~\cite{shamsian2021personalized} proposes pFedHN and only explores generating limited heterogeneous models. FLHA-GHN~\cite{Litany2022FederatedLW} uses graph hypernetworks to generate model parameters for different architectures, training the hypernetwork on local clients imposes additional computational overhead. They cannot be deployed in resource-constrained environments; in contrast, our efficient methods, MH-pFedHN and MH-pFedHNGD, demonstrate strong potential for deployment in practical scenarios.

\subsection{Model-Heterogeneous Personalized Federated Learning Methods}

Due to resource constraints~\cite{chai2020tifl,
shin2024effective}, communication  limitations~\cite{Caldas2018ExpandingTR, 10.5555/3495724.3496904, Shah2021ModelCF}, communication limitations, and personalized requirements for models~\cite{li2019fedmd,zhu2021data,wu2024exploring}, MH-pFL has become an important research direction. Even when clients share the same model architecture, two forms of model heterogeneity can still arise: 1) Due to resource differences, clients may need to employ different sub-models for training~\cite{diao2021heterofl}; 2) Clients with varying personalization requirements retain parameters of specific layers solely for local updates~\cite{li2021fedbn}, resulting in model heterogeneity. Furthermore, the model architecture may impact model privacy~\cite{10.5555/3698900.3699280}, clients are often reluctant to further disclose the details of their model design information.

\textbf{Partial Training} is adopted in some methods, where each client selects a sub-model that originates with the global model. 
The parameters from the client sub-models are aggregated by averaging the corresponding parameters. HeteroFL~\cite{diao2021heterofl} divides the global model into different sub-models, allowing clients to train appropriate models based on computational capabilities, enabling low-resource clients to contribute to the global model without being excluded from the aggregation. FedRolex~\cite{alam2022fedrolex} employs a rolling scheme that allows for sub-model extraction and for evenly training different parts of the global model on the server. DepthFL~\cite{kim2023depthfl}  utilizes different depths of different local clients and prunes the deepest layers off global model to help allocate the server model to all the clients based on computational resources. 
pFedGate~\cite{Chen2023EfficientPF} explores to learn sparse local model adaptively.
HypeMeFed~\cite{shin2024effective} combines a multi-path network architecture with weight generation to support client heterogeneity.
However, due to differences in model architecture, the parameters of different sub-models learning by partial training may not align, resulting in suboptimal performance.

\textbf{Knowledge Distillation}~\cite{hinton2015distilling} is a model compression technique, initially introduced to lower communication overhead and could reduce the impact of data heterogeneity on performance in FL~\cite{sattler2020communication,DBLP:journals/corr/abs-2108-13323}. Federated distillation~\cite{jeong2018communication}  addresses model heterogeneity when facilitating collaboration between clients with different architectures by transferring knowledge from complex models to simpler ones. 

FedMD~\cite{li2019fedmd} and DS-FL~\cite{itahara2021distillation} use a pre-existing public dataset for knowledge extraction, aggregating local soft predictions on the server. FedDF~\cite{lin2020ensemble} adopts an ensemble distillation approach to train heterogeneous models on a centralized server. KT-pFL~\cite{zhang2021parameterized} trains personalized soft prediction weights on the server to improve heterogeneous models' performance further. However, federated distillation necessitates the server to maintain a balanced shared dataset, which may be infeasible in privacy-sensitive scenarios. Inspired by data-free knowledge distillation~\cite{zhang2022dense,dai2024enhancing}, some methods~\cite{zhu2021data, zhang2022fedzkt, Zhang2023TowardsDK} use GAN~\cite{goodfellow2020generative} to generate synthetic data on the server side. DFRD~\cite{10.5555/3666122.3666906} and FedGD~\cite{Zhang2023TowardsDK} design a distributed GAN between the server and the clients to enable data-free knowledge transfer and effectively tackle model heterogeneity. However, training the key generator in GAN-based federated distillation poses significant challenges that can impact the overall performance of the system. FedAKT~\cite{liu2025adapter} combines knowledge distillation and model decoupling without the use of public data,  however, homogeneous adapters can increase computational and communication overhead for clients, and distillation and dual-head mechanisms can perform poorly in highly heterogeneous environments. Therefore, federated distillation still faces the challenges related to privacy risks and the quality of the public and synthetic datasets. 

\textbf{Model Decoupling} is another attempt from researchers, which divides the client model into feature extractors and classifier heads. Further, scientists extend the previous model decoupling methods to share part of the global model while retaining the other part in a heterogeneous form locally. For example, 

FedClassAvg~\cite{10.1145/3545008.3545073} aggregates classifier weights to establish a consensus on decision boundaries in the feature space, enabling clients with non-IID data to learn from scarce labels;
FedGH~\cite{10.1145/3581783.3611781} trains a shared, generalized global prediction head using representations extracted by clients' heterogeneous feature extractors on the FL server. In pFedES~\cite{yi2023pfedes}, the clients are trained via their proposed iterative learning method to facilitate the exchange of global knowledge; small local homogeneous extractors are then uploaded for aggregation to help the server learn the knowledge across clients.
However, these basic mechanisms might find it difficult to achieve effective knowledge fusion in complex heterogeneous scenarios, which results in limitations in final performance. 

\begin{figure}[t]
  \centering
  \begin{minipage}[t]{0.5\textwidth}
    \vspace{0pt} 
    \centering
    \includegraphics[width=\linewidth]{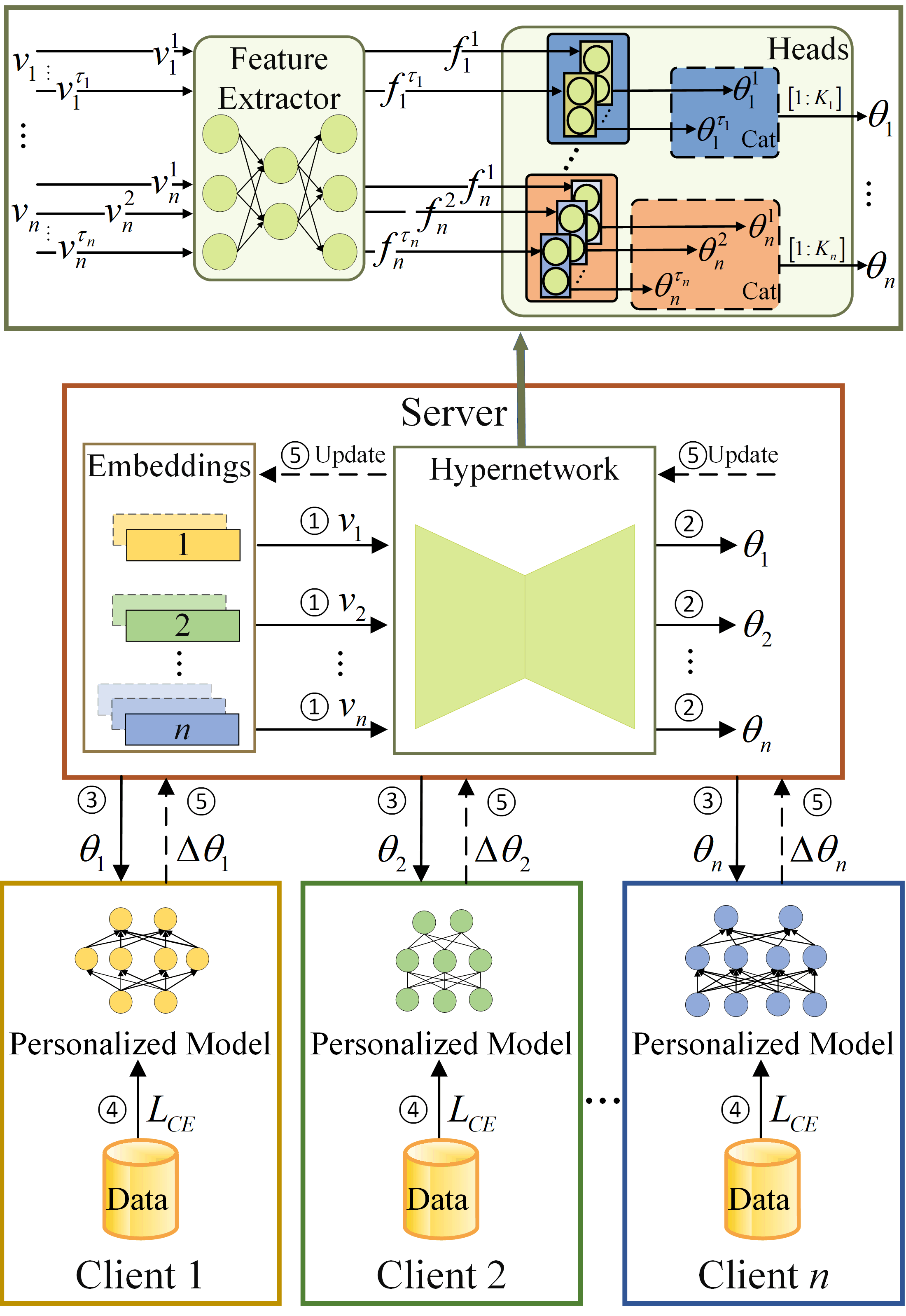} 
    \caption{Framework of MH-pFedHN. The workflow contains 5 steps: \ding{172} input the embedding matrix $\bm{v}_i$ of client $i$ into the hypernetwork; \ding{173} generates the parameters $\bm{\theta}_i$ of client $i$; \ding{174} client $i$ receives the parameters $\bm{\theta}_i$ from the server; \ding{175} client $i$ trains the personalized model $\bm{\theta}_i$ using the private data; \ding{176} client $i$ uploads the update of parameters $\Delta\bm{\theta}_{i}$ to the server; server updates customized embedding vectors and hypernetworks.}
    \label{fig:pic2}
  \end{minipage}
  \hfill
  \begin{minipage}[t]{0.47\textwidth}
    \vspace{0pt} 
    \centering
    \includegraphics[width=\linewidth]{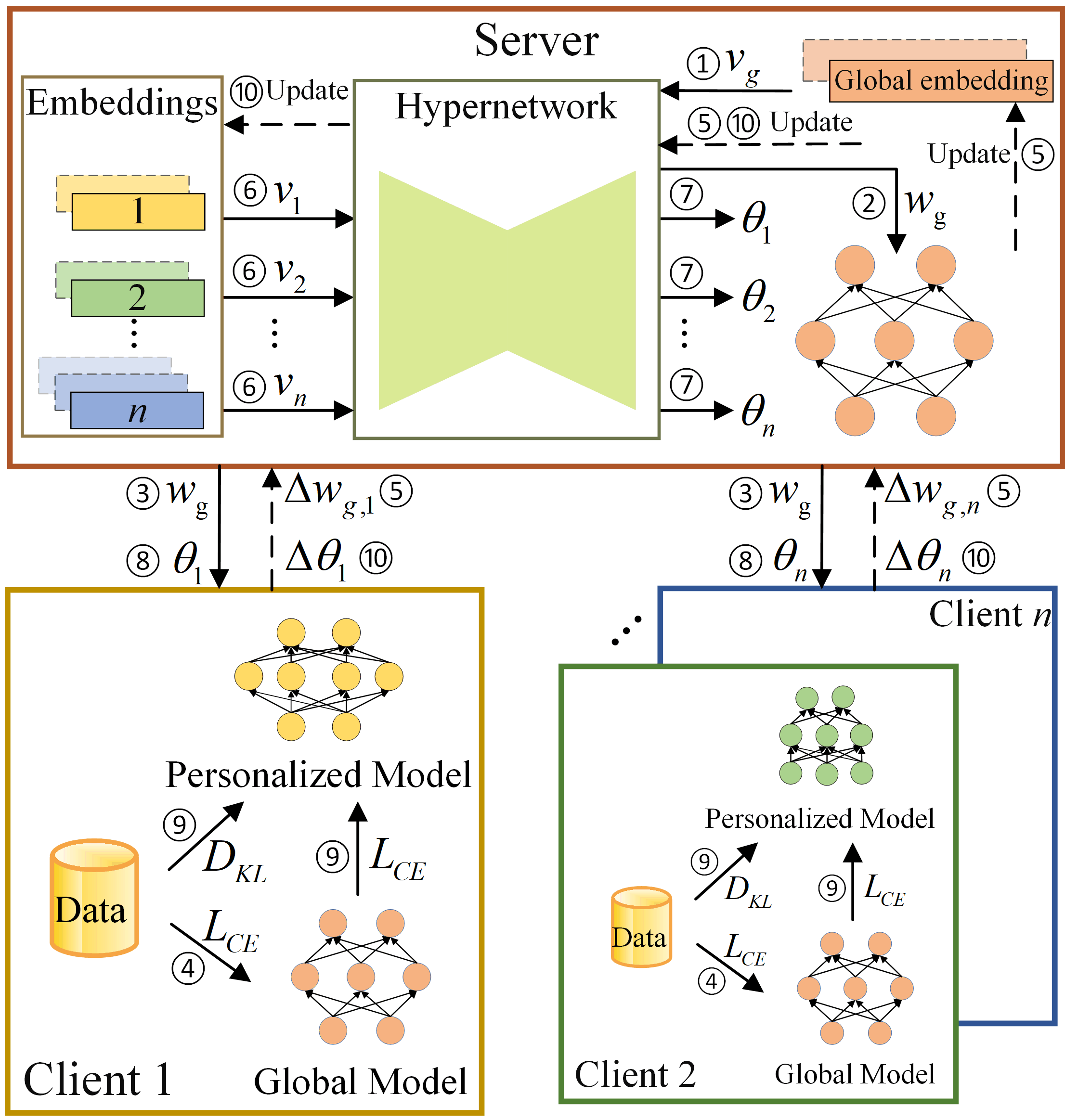} 
    \caption{Framework of MH-pFedHNGD. The workflow contains 10 steps: \ding{172} input the embedding matrix $\bm{v}_g$ of the global model into the hypernetwork; \ding{173} generates the parameters $\bm{w}_g$ of the global model; \ding{174} each client receives the global model parameters $\bm{w}_g$ from the server; \ding{175} each client trains the global model using private data; \ding{176} The server receives $\Delta \bm{w}_{g,i}$ from each client and updates the hypernetwork and global embedding vector parameters; \ding{177} input the embedding matrix $\bm{v}_i$ of client $i$ into the hypernetwork; \ding{178} generates the parameters $\bm{\theta}_i$ of client $i$; \ding{179} client $i$ receives the parameters $\bm{\theta}_i$ from the server; \ding{180} the global model $\bm{w}_g$ assists in training personalized model $\bm{\theta}_i$ of client $i$ on private data; \ding{181} client $i$ uploads update of parameters $\Delta\bm{\theta}_{i}$ to the server; server updates customized embedding vectors and hypernetworks.}
    \label{fig:pic3}
  \end{minipage}
\end{figure}

\section{Our Framework}
In this section, we first formalize the MH-pFL problem. 
Then, we introduce MH-pFedHN, which generates parameters for different models through customized embedding vectors and utilizes a hypernetwork for knowledge fusion across heterogeneous clients.
Finally, we present MH-pFedHNGD, which utilizes a lightweight global model to enhance the learning and generalization capabilities of the hypernetwork. The lightweight plug-in component also assists personalized model training for clients via knowledge distillation, further improving model training across heterogeneous clients. 
The overviews of our two methods are shown in Figures~\ref{fig:pic2} and \ref{fig:pic3}, where one round of MH-pFedHN consists of 5 steps and one round of MH-pFedHNGD is made up of 10 steps.

\subsection{Problem Formulation}
Our goal is to develop an MH-pFL approach that, without requiring knowledge of model architectures, meets the specific requirements of each client to address the challenges brought about by the heterogeneity of the model and data.
This can be turned into the following minimization problem, which aims to capture the personalized objectives of each client while accommodating heterogeneous data distributions and model architectures
\begin{equation}
        \label{equation1}
	\{\bm{\theta}^*_{1},\dots,\bm{\theta}^*_{n}\} =\underset{\bm{\theta_{1}},\dots,\bm{\theta}_{n}}{\operatorname {arg\,min}}{ \sum_{i=1}^n E_{x,y \sim P_i}\left[\ell_i(x_j, y_j; \bm{\theta}_{i})\right]},
\end{equation}
where \(P_i\) represents the local data distribution of the \(i\)-th client, and \(\ell_i(x_j, y_j;\bm{\theta}_i)\) denotes the loss function for the \(i\)-th client's model with parameters \(\bm{\theta}_i\). In particular, \(\bm{\theta}_i\) is specific to each client and can correspond to models of varying architectures.

To further formalize the optimization problem, the training objective is expressed as  
\begin{equation}
	\label{equation2}
	\underset{\bm{\theta}_{1},\dots,\bm{\theta}_{n}}{\operatorname {arg\,min}} {\sum_{i=1}^n L_i(\bm{\theta}_i)} = \underset{\bm{\theta}_{1},\dots,\bm{\theta}_{n}}{\operatorname {arg\,min}} \sum_{i=1}^n \frac{1}{m_i} \sum_{j=1}^{m_i} \ell_i(x_j, y_j; \bm{\theta}_{i}),
\end{equation}
where \(m_i\) denotes the number of data samples in the local dataset of the \(i\)-th client, and $L_i(\bm{\theta}_i)$ represents the empirical loss over its dataset. By allowing each client to optimize its model, ensuring that the learned parameters $\bm{\theta}_i$ are suited to the client's unique data distribution and task.

\subsection{MH-pFedHN}

Here, we describe MH-pFedHN to the MH-pFL problem (Equation~\ref{equation2}), and the complete algorithm for MH-pFedHN is provided in Algorithm~\ref{alg1}. Let \( h(\cdot; \bm{\varphi}) \) denote the hypernetwork $h$ with parameters \( \bm{\varphi} \), where \( \bm{\varphi} \) is composed by a feature extractor  \(\bm{\varphi}_f\) and multiple heads \{\(\bm{\varphi}_{H_l} \)\}. 

In the initial stage, clients need to upload the number of parameters $K$ required for their personalized heterogeneous models. The server then dynamically determines the hypernetwork output dimension \( N \) based on \( K \) (we recommend that the value of \(\left\lceil K_i / N \right\rceil\) should be greater than the number of layers in the client model) and calculates the customized number of embedding vectors needed for $i$-th client as \(\tau_i = \left\lceil K_i / N \right\rceil\).  Therefore, models with similar parameters share the same customized embedding vectors. For these personalized heterogeneous models, a shared head \(\bm{\varphi}_{H_l}\) is established.

\textit{Specifically}, for example, if two models have the same number of parameters and are both determined to have three customized embedding vectors, they will share the same head \(\bm{\varphi}_{H_l}\). This head has three output channels (the channel number equals to the number of customized embedding vectors), each tailored for a specific customized embedding vector input. The shared feature extractor for all clients and this head  \(\bm{\varphi}_{H_l}\) will generate three subsets of parameters with length \( N \) for both models. These three subsets of parameters, when combined and rounded down according to the specific parameter size of each client, constitute the personalized parameters for each client. \textit{Generally}, for the \( j \)-th customized embedding vector of client \( i \) (associated with the \( l \)-th head), the output from the feature extractor is processed as \(
\bm{\theta}_i^j = h(\bm{v}_i^j; \bm{\varphi}_f, \bm{\varphi}_{H_l})
\), where we use $\bm{v}_{i} = [\bm{v}^{1}_{i},\dots,\bm{v}^{\tau_i}_{i}]$ to denotes the customized embedding vectors for the $i$-th client $\bm{\theta}_i$ = $\bm{\theta}_{i}(\bm{\varphi}) := h(\bm{v}_i; \bm{\varphi})_{[1:K_{i}]}$. \textit{Finally}, the personalized model parameters for client \( i \) are generated as follows:
\begin{equation}
\label{eq:thetacombine}
\bm{\theta}_i := \text{concat}(\bm{\theta}_i^1, \bm{\theta}_i^2, \cdots, \bm{\theta}_i^{\tau_i})_{[1:K_i]},
\end{equation}
After client $i$ trains the personalized model based on their private data, the hypernetwork $h$ updates a set of $\Delta\bm{\theta}_i$. Therefore, our MH-pFedHN allows the server to generate parameters for multiple models with a similar number of parameters in a single pass.

Based on the aforementioned setup, we adopt the MH-pFL objective (Equation~\ref{equation2}) and get our MH-pFedHN optimization function as:
\begin{equation}
\label{equation3}
\begin{aligned}  
   \underset{\bm{\varphi}, \bm{v}_1, \dots, \bm{v}_n}{\operatorname {arg\,min}} \sum_{i=1}^n L_i \left(h(\bm{v}_i; \bm{\varphi})_{[1:K_{i}]}\right) 
    &=\underset{\bm{\varphi}, \bm{v}_1, \dots, \bm{v}_n}{\operatorname {arg\,min}} \sum_{i=1}^n \frac{1}{m_i} \sum_{j=1}^{m_i}  \left(\ell_i(x_j, y_j;  \left(h(\bm{v}_i; \bm{\varphi})_{[1:K_{i}]}\right)\right) \\
    & =\underset{\bm{\theta}_{1},\dots,\bm{\theta}_{n}}{\operatorname {arg\,min}} \sum_{i=1}^n \frac{1}{m_i} \sum_{j=1}^{m_i} \ell_i(x_j, y_j; \bm{\theta}_{i}),
    \end{aligned}
\end{equation}

\subsection{MH-pFedHNGD}
To enhance the hypernetwork's learning capacity and generalization ability while enabling clients to efficiently extract knowledge representations from a global model, we propose MH-pFedHNGD, which introduces a lightweight global model based on MH-pFedHN. This design aims to improve overall performance at the cost of minimal system and communication overhead. The complete algorithm for MH-pFedHNGD can be found in Algorithm~\ref{alg2}. 

Thus, the global model's number of parameters is set as $K_g = \min\{K_1,\cdots, K_n\}$, thus the global model shares the same head with the client having the fewest number of parameters to make sure the global model lightweight. Then the number of global customized embedding vectors is set as \(\tau_g = \left\lceil K_g / N \right\rceil\), the global embedding vector $\bm{v}_{g} = [\bm{v}^{1}_{g},\dots,\bm{v}^{\tau_g}_{g}]$. Assuming the client with the fewest number of parameters corresponds to the $l$-th head, the global model parameters are generated by:
\begin{equation}
\label{eq:globalgeneration}
 \bm{w}_g:=h(\bm{v}_g; \bm{\varphi})_{[1:K_{g}]}=\text{concat}
 \left(
 h(\bm{v}_g^1; \bm{\varphi}_f, \bm{\varphi}_{H_{k_l}}),h(\bm{v}_g^2; \bm{\varphi}_f, \bm{\varphi}_{H_{k_l}}), \cdots, h(\bm{v}_g^{\tau_g}; \bm{\varphi}_f, \bm{\varphi}_{H_{k_l}})\right)_{[1:K_g]},
\end{equation}

Then, each client receives the parameters $\bm{w}_{g,i}=\bm{w}_g$ and trains on private data, which is to optimize: 
\begin{equation}
 \underset{\bm{\varphi}, \bm{v}_g}{\operatorname {arg\,min}} \sum_{i=1}^n \frac{m_i}{M}L_i \left(h(\bm{v}_g; \bm{\varphi})_{[1:K_{g}]}\right)=\underset{w_i}{\operatorname {arg\,min}} \sum_{i=1}^{n}\frac{m_i}{M}
    L_i(\bm{w}_{g,i}), 
\end{equation}
\begin{equation}
L_i(\bm{w}_{g,i})=\frac{1}{m_i} \sum_{j=1}^{m_i} \ell_i(x_j, y_j; \bm{w}_{g,i}),  
\end{equation}

where $M=\sum_i m_i$. After training its own global model completed in Step\ding{175} in Figure~\ref{fig:pic3}, clients upload $\Delta \bm{w}_{g,i} = \bm{w}_{g,i} - \bm{w}_g$ to server to update the hypernetwork one more time along with global customized embedding vectors, which enhances the learning and generalization capabilities of the hypernetwork.

Finally, in the distillation training phase, client $i$ reloads the global model in Step\ding{174} and receives the parameters $\bm{\theta}_i$ of the personalized model, utilizes the global model serves as a teacher model, directing the training of the personalized model for client $i$. We adopt the optimization problem (Equation~\ref{equation3}) and get our MH-pFedHNG optimization function as follows:

\begin{equation}  
    \begin{aligned}  
       \underset{\bm{\varphi}, \bm{v}_1, \dots, \bm{v}_n}{\operatorname{arg\,min}} \sum_{i=1}^n \Bigg[   
        &\lambda L_i \left( h(\bm{v}_i; \bm{\varphi})_{[1:K_{i}]} \right) +(1 - \lambda) L_{KL} \left( h(\bm{v}_i; \bm{\varphi})_{[1:K_{i}]}, h(\bm{v}_g; \bm{\varphi})_{[1:K_{g}]} \right)   
        \Bigg],
    \end{aligned}  
\end{equation}
where $ L_{KL} $ is the Kullback–Leibler divergence~\cite{Csiszr1975IDivergenceGO}, $\lambda$ is a hyperparameter used to balance the distillation loss and the cross-entropy loss. Therefore, the lightweight plug-in component further improves model training across heterogeneous clients.

\section{Experiments}
\subsection{Experiment Setup}
\textbf{Datasets.}  
We evaluate the MH-pFedHN and MH-pFedHNGD framework over four datasets, EMNIST~\cite{CohenATS17}, CIFAR-10~\cite{2009Learning}\footnote{The experiments on CIFAR-10 are in Appendix~\ref{append:cifar10}.}, CIFAR-100~\cite{2009Learning}, and Tiny-ImageNet~\cite{le2015tiny}. We adopt two non-IID settings~\cite{t2020personalized,lin2020ensemble,
li2022federated,liu2024fedlpa}. 
1) Quantity-based label imbalance (non-IID\_1). 
For the EMNIST, CIFAR-100, and Tiny-ImageNet datasets, we randomly allocate 6, 10, and 20 classes to each client, respectively.  We draw \( \alpha_{i,c} \sim U(0.4, 0.6) \), and allocate \( \frac{\alpha_{i,c}}{\sum_{j} \alpha_{j,c}} \) of the samples for the class \( c \) selected on client \( i \). 2) Distribution-based label imbalance (non-IID\_2). We employ the Dirichlet distribution \( Dir(0.01) \) to partition the dataset among the clients. For each client, 75\% of the data is used for training, and 25\% is used for testing. As the EMNIST is simpler than others, we only report results for 200 clients.

\textbf{Models.}

In prior pFL studies, each client employs a LeNet-style model~\cite{lecun1998gradient}. For a fair comparison, we use this model in the homogeneous model experiments. In addition, we use a  VGGNet~\cite{simonyan2014very}, along with three residual networks~\cite{He_2016_CVPR, zagoruyko2016wide}. All of our heterogeneous experiments use these five models, which are evenly distributed among all clients by default. The feature extractor of hypernetwork is comprised by a three-layer fully connected network. The details are in Appendix~\ref{append:ModelArchitectures}.

\textbf{Baselines.}
We choose various state-of-the-art methods. \textbf{pFedHN} ~\cite{shamsian2021personalized} introduces hypernetwork to directly produce personalized model; \textbf{pFedLA}~\cite{ma2022layer} utilizes hypernetwork to compute aggregation weights for the local models of each client; \textbf{FedGH}~\cite{10.1145/3581783.3611781} uses a generalized global prediction header for diverse model structures; \textbf{pFedLHN}~\cite{zhu2023layer} leverages a layer-wise hypernetwork to achieve fine-grained personalization across different layers of the model; \textbf{PeFLL}~\cite{scott2024pefll} uses a learning-to-learn approach to generate personalized models efficiently; \textbf{FedAKT}~\cite{liu2025adapter} employs homogeneous adapters and feature distillation mechanisms to enhance knowledge transfer and model adaptation in heterogeneous environments. We also include \textbf{Local} Training without aggregation and \textbf{FedAvg}~\cite{mcmahan2017communication}.

\textbf{Training Strategies.}
We have at most 500 server-client communication rounds, with results averaged over three runs. For MH-pFedHN, we set local epochs to 2, SGD optimizer with the learning rate \(1\mathrm{e}{-3}\) and weight decay \(1\mathrm{e}{-4}\) and momentum 0.9, batch size 64. The hypernetwork uses the Adam optimizer~\cite{Kingma2014AdamAM} with a learning rate of \(2\mathrm{e}{-4}\),  embedding dimension 64, and the output dimension 3072. For MH-pFedHNGD, we configure the global model LeNet-5. On CIFAR-100, Tiny-ImageNet, and EMNIST datasets, we set distillation temperatures as 15, 24, and 10, and distillation loss coefficients as 0.01, 0.2, and 0.1, respectively. More settings and design choices are in Appendix~\ref{append:expsetting} and \ref{append:designlabels}.

\begin{table*}[!t]  
        \belowrulesep= 0pt
        \aboverulesep= 0pt
	\centering  
	\begin{minipage}{\textwidth}  
		\caption{Homogeneous model experiments, where left side is non-IID\_1 and right is non-IID\_2. Bold indicates that our method outperforms all the baselines.}  
		\label{table_ho_1_and_table_ho_2}  
		\begin{small}  
			\begin{sc}  
                \scalebox{0.68}{
				\begin{tabular}{c|c|c|c|c|c|c|c}  
					\toprule  
					& \multicolumn{3}{c|}{CIFAR-100} & \multicolumn{3}{c|}{\textnormal{Tiny-ImageNet}} & \multicolumn{1}{c}{EMNIST}  \\   
					\cmidrule(){2-8} 
					\# \textnormal{Clients} & 50 & 100 & 200 & 50 & 100 & 200 & 200\\   
					\midrule  
					\textnormal{Local} & 50.32 \quad 72.79 & 40.41 \quad 73.52 & 34.65 \quad 73.56  & 29.17 \quad 50.65 & 20.73 \quad 55.14 & 14.90 \quad 55.81 & 96.26 \quad 98.70   \\
					\textnormal{FedAvg~\cite{mcmahan2017communication}} & 22.94 \quad 22.59 & 24.80 \quad 24.27 & 25.55 \quad 23.07 & 8.13 \quad 6.64 & 8.80 \quad 7.54 & 9.12 \quad 8.22 & 81.49 \quad 80.18  \\ \midrule 
				\textnormal{pFedHN~\cite{shamsian2021personalized}} & 63.66 \quad 76.76 & 58.90 \quad 79.74 & 32.77 \quad 74.14 & 40.10 \quad 56.76 & 35.39 \quad 58.93  & 32.10 \quad 61.12 & 97.50 \quad 99.18 \\
                    	\textnormal{pFedLA~\cite{ma2022layer}} & 63.33 \quad 72.86 & 55.83 \quad 75.22 & 55.36 \quad 75.88 & 39.69 \quad 48.30 & 30.00 \quad 53.59  &23.42 \quad 54.84 &95.41 \quad 98.40 \\
                        \textnormal{FedGH~\cite{10.1145/3581783.3611781}} & 61.01 \quad 75.54 & 53.61 \quad 76.65 & 38.70 \quad 77.30 & 31.67 \quad 54.30 & 23.64 \quad 57.97  & 17.80 \quad 57.44 & 96.17 \quad 96.42 \\
					\textnormal{pFedLHN~\cite{zhu2023layer}}  & 61.00 \quad 77.03& 58.39 \quad 79.06 & 53.43 \quad 79.49 & 39.49 \quad 55.71 & 35.00 \quad 60.62  & 31.31 \quad 58.30 & 97.52 \quad \textbf{99.25} \\
                    \textnormal{PeFLL~\cite{scott2024pefll}}  & 50.64 \quad 66.79 & 49.20 \quad 71.95 & 40.97 \quad 73.07 & 32.41 \quad 51.44 & 28.07 \quad 55.86 &23.65 \quad 56.26 &94.68 \quad 98.88  \\ 
                    \textnormal{FedAKT~\cite{liu2025adapter}} & 60.85 \quad 74.28 & 56.09 \quad 77.92 & 51.48 \quad 78.56 & 38.57 \quad 55.23 & 31.86 \quad 60.23  & 31.87 \quad 61.58 & 97.21 \quad 98.96 \\ \midrule  
					\textnormal{MH-pFedHN (ours)} & \textbf{64.69} \quad \textbf{77.93} & \textbf{63.32} \quad \textbf{80.93} & \textbf{60.11} \quad \textbf{81.60}  & \textbf{42.35} \quad \textbf{58.30} & \textbf{37.62} \quad \textbf{62.68} & \textbf{37.30} \quad \textbf{63.61} & \textbf{97.56} \quad 99.16  \\
					\textnormal{MH-pFedHNGD (ours)} &\textbf{68.30} \quad \textbf{80.07} &\textbf{63.97} \quad \textbf{82.51} &\textbf{61.59} \quad \textbf{82.54} & \textbf{44.44} \quad \textbf{58.35} & \textbf{40.99} \quad \textbf{63.39} & \textbf{39.96} \quad \textbf{66.67} & \textbf{97.76} \quad 98.83  \\
					\bottomrule  
				\end{tabular}  
                }
			\end{sc}  
		\end{small}  
	\end{minipage}  
    \vspace{-0.1cm} 
\end{table*}

\subsection{Homogeneous Model Experiments}
\label{Homogeneous Model}

The results in Table~\ref{table_ho_1_and_table_ho_2} demonstrate that our proposed methods, MH-pFedHN and MH-pFedHNGD, both outperform all the baseline approaches under both non-IID scenarios. Although the accuracy of all methods decreases as the number of clients increases, our methods consistently maintain high accuracy, underscoring their robustness in handling data heterogeneity and scalability to many clients. MH-pFedHNGD achieves consistently higher test accuracy compared to MH-pFedHN, further validating the effectiveness of generating a global model for the hypernetwork update and leveraging it for distillation training. Furthermore, most methods show high test accuracy (up to 96\%) on EMNIST dataset due to its simplicity. However, this feature limits observable performance differences, making meaningful comparisons challenging on EMNIST dataset. In contrast, our method is better equipped to handle complex real-world data scenarios.

\begin{table*}[!t]  
	\centering  
        \belowrulesep= 0pt
        \aboverulesep= 0pt
	\begin{minipage}{\textwidth}  
		\caption{Heterogeneous model experiments, where left side is non-IID\_1 and right is non-IID\_2.}  
		\label{table_he_1_and_table_he_2}  
		\begin{small}  
			\begin{sc}  
            \scalebox{0.68}{
				\begin{tabular}{c|c|c|c|c|c|c|c}  
					\toprule  
					& \multicolumn{3}{c}{CIFAR-100} & \multicolumn{3}{c}{\textnormal{Tiny-ImageNet}} & \multicolumn{1}{c}{EMNIST}  \\   
					\cmidrule(){2-8}   
					\# \textnormal{Clients} & 50 & 100 & 200 & 50 & 100 & 200 & 200\\   
					\midrule  
					\textnormal{Local} & 50.74 \quad 71.69 & 40.41 \quad 72.01 & 31.95 \quad 73.05 & 29.64 \quad 51.64 & 19.63 \quad 54.05 & 14.44 \quad 53.92  & 96.77 \quad 98.83  \\
					\textnormal{FedAvg~\cite{mcmahan2017communication}}  & 22.80 \quad 17.27 & 24.72 \quad 22.18 & 24.28 \quad 21.29 & 8.17 \quad 8.22 & 8.59 \quad 11.86 & 9.11 \quad 16.40  & 85.09 \quad 84.78  \\ \midrule  

					\textnormal{pFedHN~\cite{shamsian2021personalized}} & 50.93 \quad 73.22 & 42.82 \quad 74.73 & 12.49 \quad 72.62 & 32.15 \quad 54.29 & 24.33 \quad 58.31 & 15.64 \quad 57.82 & 97.39 \quad 99.07  \\
                    \textnormal{pFedLA~\cite{ma2022layer}} & 51.97 \quad 71.91 & 52.06 \quad 74.68 & 40.11 \quad 75.24 & 26.97 \quad 46.86 & 19.70 \quad 51.73 & 17.28 \quad 54.44 & 93.82 \quad 98.36  \\
                    \textnormal{FedGH~\cite{10.1145/3581783.3611781}} & 52.61 \quad 71.38 & 42.45 \quad 73.25 & 31.57 \quad 71.57 & 25.69 \quad 50.73 & 16.90 \quad 53.57  & 11.74 \quad 55.04 & 96.01 \quad 96.37 \\
	
                    \textnormal{pFedLHN~\cite{zhu2023layer}} & 55.91 \quad 75.28 & 48.46 \quad 76.16 & 40.96 \quad 74.60 & 38.13 \quad 55.13 & 29.20 \quad 59.36 & 20.85 \quad 57.53  & 97.47 \quad 99.15 \\ 
        
                    \textnormal{PeFLL~\cite{scott2024pefll}}  &55.87 \quad 72.28 & 52.09 \quad 74.51 & 42.56 \quad 72.79 & 35.70 \quad 55.21 & 30.08 \quad 55.60 & 25.24 \quad 55.01 & 97.46 \quad 99.10  \\ 
                    \textnormal{FedAKT~\cite{liu2025adapter}} & 53.28 \quad 73.15 & 43.80 \quad 75.25 & 35.37 \quad 74.98 & 35.57 \quad 53.74 & 28.25 \quad 56.95  & 20.17 \quad 55.36 & 97.43 \quad 99.11 \\ \midrule  
					\textnormal{MH-pFedHN (ours)} & \textbf{57.09} \quad \textbf{75.40} & 50.35 \quad \textbf{77.24} & \textbf{42.98} \quad \textbf{75.38} & 38.00 \quad \textbf{55.51} & \textbf{30.32} \quad 58.93 & 23.31 \quad \textbf{58.79} & \textbf{97.58} \quad \textbf{99.18}   \\
					\textnormal{MH-pFedHNGD (ours)} & \textbf{60.11} \quad \textbf{76.76} & \textbf{52.14} \quad 77.03 & \textbf{43.41} \quad \textbf{76.46} & \textbf{42.18} \quad \textbf{58.02} & \textbf{34.98} \quad \textbf{61.11} & \textbf{26.12} \quad \textbf{60.38} & \textbf{97.65} \quad 98.73 \\
					\bottomrule  
				\end{tabular}  
            }
			\end{sc}  
		\end{small}  
	\end{minipage}  
    \vspace{-0.2cm} 
\end{table*}

\subsection{Heterogeneous Model Experiments}
\label{Heterogeneous Model}

For pFedHN and pFedLHN, we modify them by adding multiple heads, enabling them to generate parameters for different models for heterogeneous model experiments. 
As PeFLL, pFedLA, FedAvg, and Local Training do not allow model heterogeneity, we thus conduct repeat experiments using various models and take the average test accuracy as the final results.

Experiment results in Table~\ref{table_he_1_and_table_he_2} indicate that methods that allow model heterogeneity show a decline in accuracy under the same settings in most cases for homogeneous models. This suggests that collaboration among heterogeneous clients is still insufficient and challenging, an issue we need to address in the future.  For PeFLL, pFedLA, FedAvg, and Local Training that do not allow model heterogeneity, the results are average based on LeNet, VGGNet, and ResNet, which may be better than the results in Table~\ref{table_ho_1_and_table_ho_2} as they only have LeNet. Overall, our two approaches still outperform all the baseline methods. Furthermore, MH-pFedHNGD shows significant improvement over MH-pFedHN in most settings, highlighting the necessity of introducing a global model and conducting knowledge distillation training in settings with model heterogeneity.

\subsection{Experiment with Generalization}
Generalization in FL refers to the model's ability to perform well on unseen clients. This section evaluates the generalization capabilities of MH-pFedHN and MH-pFedHNGD. These experiments utilize the CIFAR-100 dataset with 100 clients, of which 80 clients are used to train the hypernetwork, while the remaining 20 clients are designated for generalization testing. During the testing phase, the parameters of the hypernetwork remain fixed, and only the customized embedding vectors corresponding to each client are optimized.

\begin{figure}[t]
    \begin{minipage}{0.485\textwidth}
        \centerline{\includegraphics[scale=0.17]{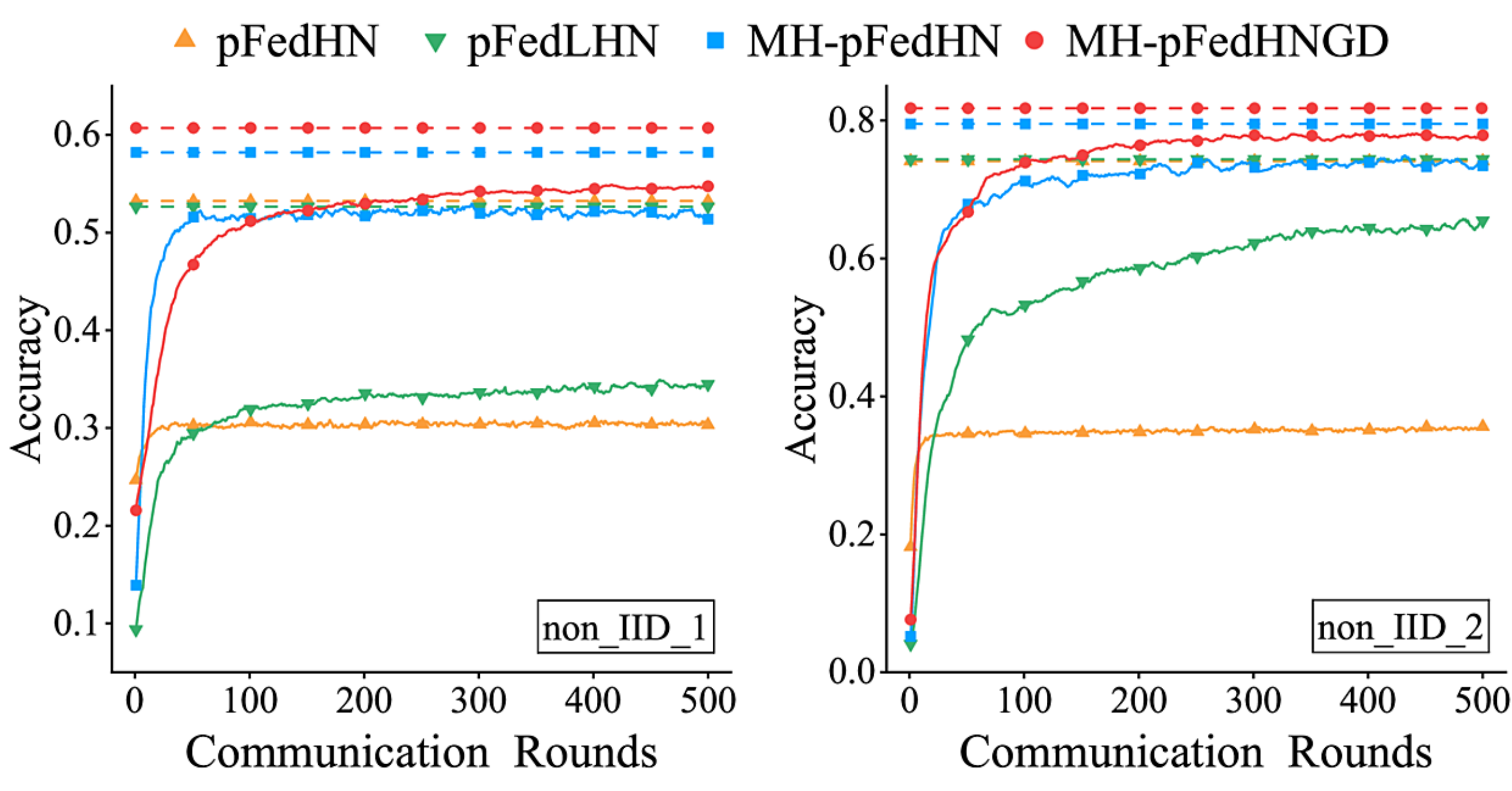}}
        \caption{Generalization experiments in the homogeneous scenario, where solid lines denote test clients and dashed lines denote training clients.}
        \label{cnn_gen1_and_2}
    \end{minipage}
    \hfill
    \begin{minipage}{0.485\textwidth}
        \centerline{\includegraphics[scale=0.17]{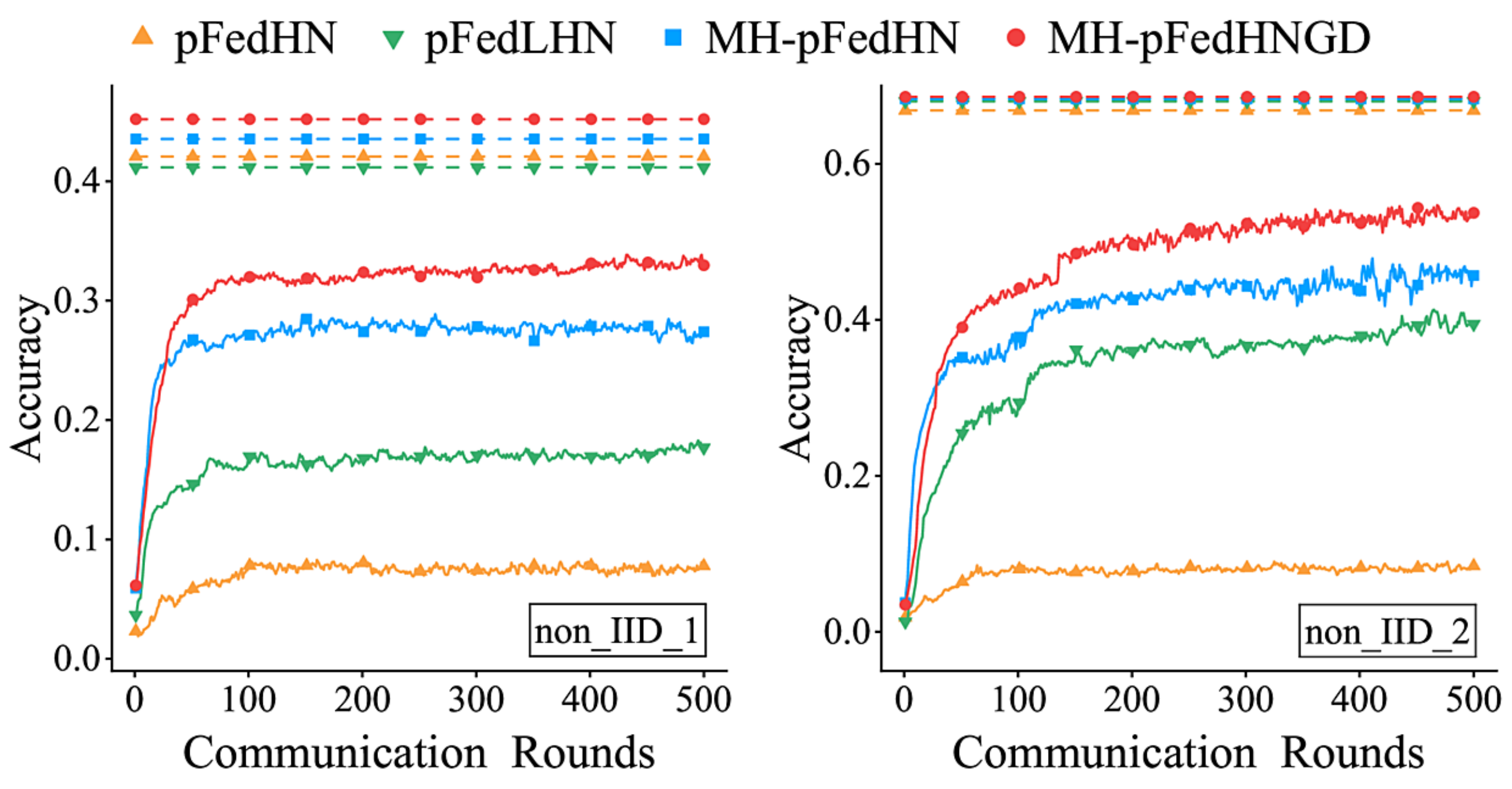}}
        \caption{Generalization experiments in the heterogeneous scenario, where solid lines denote test clients and dashed lines denote training clients.}
        \label{more_gen1_and_2}
    \end{minipage}
\end{figure}

\textbf{Generalize to homogeneous clients.}
Figure~\ref{cnn_gen1_and_2} shows that MH-pFedHNGD and MH-pFedHN exhibit outstanding generalization capabilities, significantly outperforming other baseline methods up to 20\%. In the early stages of generalization, both algorithms converge rapidly and achieve accuracy on the test set comparable to that of the training clients. Moreover, the generalization ability of MH-pFedHNGD is superior to that of MH-pFedHN, indicating that introducing a global model is beneficial for further enhancing the generalization performance of the hypernetwork.

\textbf{Generalize to heterogeneous clients.}
In the experiments with heterogeneous models, we used the LeNet, VGGNet, and ResNet. We then added the MLP model and SqueezeNet1\_0 model~\cite{iandola2016squeezenet}. These five models were evenly distributed between clients. Figure~\ref{more_gen1_and_2} shows that our method still exhibits the better generalization performance (up to 50\%) of all the baselines for heterogeneous models, significantly outperforming the other two baselines. Again, the generalization ability of MH-pFedHNGD is notably superior to MH-pFedHN.

\begin{figure}[ht]
  \begin{minipage}[t]{0.48\textwidth}
    \begin{minipage}[t]{\textwidth}
      \centering
      \vspace{-0.2cm}
      \includegraphics[scale=0.163]{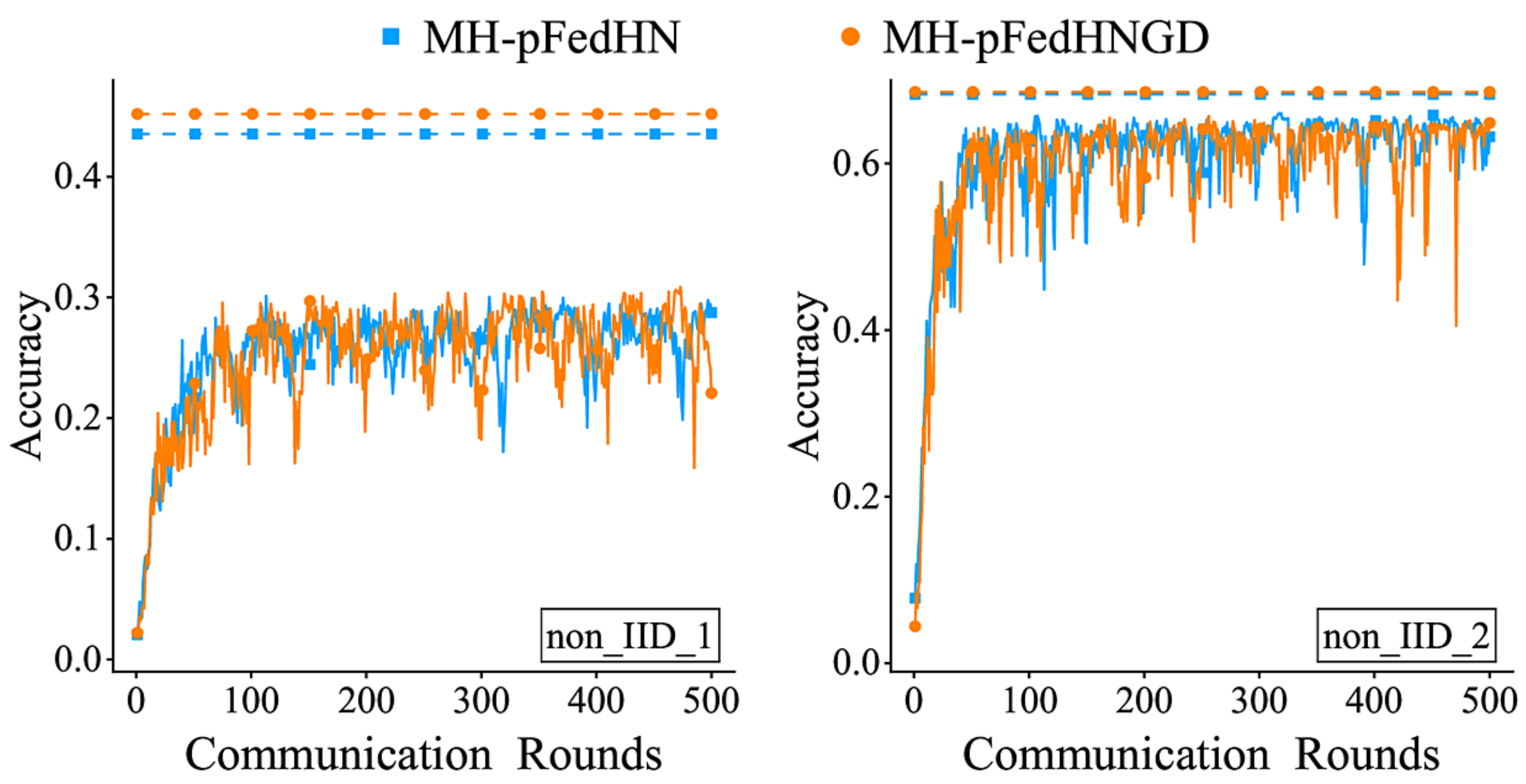}
      \caption{Generalization experiments for new architectures, where solid lines denote test clients and dashed lines denote training clients.}
      \label{more_new}
    \end{minipage}
    \vspace{-0.1cm}
    
    \begin{minipage}[t]{\textwidth}
      \centering
      \captionof{table}{
      The ablation experiments of MH-pFedHN. Green indicates the setting without a head, and Red indicates the setting where each client has one head. ``-'' indicates out of memory.}
      \label{table_a_MH-pFedHN}
      \vspace{-0.2cm}
      \belowrulesep=0pt
      \aboverulesep=0pt
      \begin{small}
        \begin{sc}
          \scalebox{0.68}{
            \begin{tabular}{l|ccc|ccc}  
              \toprule
              & \multicolumn{6}{c}{CIFAR-100}   \\ 
              \cmidrule(lr){2-7} 
              \textnormal{Data\_Distribution}
              & \multicolumn{3}{c|}{\textnormal{non-IID\_1}} 
              & \multicolumn{3}{c}{\textnormal{non-IID\_2}} \\
              \cmidrule(lr){2-7} 
              \# \textnormal{Clients} & 50 & 100 & 200 & 50 & 100 & 200  \\ 
              \midrule  
              \rowcolor{LightGreen}
              \textnormal{homogeneous}  &58.32 & 47.86 & 40.63 & 76.01 & 77.26 & 76.76 \\
              \rowcolor{LightGreen}
              \textnormal{heterogeneous} & 39.72& 35.40 & 29.28 & 57.74 & 64.27 & 60.68 \\
              \midrule 
              \rowcolor{LightRed}
              \textnormal{homogeneous}  & 53.51& - & - & 73.70 & - & -\\
              \rowcolor{LightRed}
              \textnormal{heterogeneous} & - & - & - & - & - & -\\
              \bottomrule
            \end{tabular}
          }
        \end{sc}
      \end{small}
      \vspace{-0.5cm}
    \end{minipage}
  \end{minipage}
  \hfill
  \begin{minipage}[t]{0.48\textwidth}
    \begin{minipage}[t]{\textwidth}
      \centering
      \captionof{table}{The ablation experiments of MH-pFedHNGD. Green and red correspond to homogeneous and heterogeneous models, respectively.}
      \label{table_a}
      \vspace{-0.2cm}
      \belowrulesep=0pt
      \aboverulesep=0pt
      \begin{small}
        \begin{sc}
          \scalebox{0.68}{
            \begin{tabular}{l|ccc|ccc}  
              \toprule
              & \multicolumn{6}{c}{CIFAR-100}   \\ 
              \cmidrule(lr){2-7} 
              \textnormal{Data\_Distribution}
              & \multicolumn{3}{c|}{\textnormal{non-IID\_1}} 
              & \multicolumn{3}{c}{\textnormal{non-IID\_2}} \\
              \cmidrule(lr){2-7} 
              \# \textnormal{Clients} & 50 & 100 & 200 & 50 & 100 & 200  \\ 
              \midrule  
              \rowcolor{LightGreen}
              \textnormal{MH-pFedHN}  &63.91 & 63.14 & 59.25 & 78.25 & 80.79 & 81.04 \\
              \rowcolor{LightGreen}
              \textnormal{MH-pFedHNG}  &66.19 & 63.36 & 59.71 & 79.18 & 81.80 & 81.85 \\
              \rowcolor{LightGreen}
              \textnormal{MH-pFedHNGD}  &68.27 & 63.58 & 61.19 & 80.18 & 82.31 & 82.19 \\
              \midrule 
              \rowcolor{LightRed}
              \textnormal{MH-pFedHN}  &57.18 & 50.35 & 40.83 & 74.84 & 76.78 & 74.71\\
              \rowcolor{LightRed}
              \textnormal{MH-pFedHNG} & 57.01 & 49.81 & 41.45 & 75.44 & 76.32 & 76.04 \\
              \rowcolor{LightRed}
              \textnormal{MH-pFedHNGD} & 60.11 & 51.57 & 43.41 & 76.76 & 77.03 & 76.46 \\
              \bottomrule
            \end{tabular}
          }
        \end{sc}
      \end{small}
    \end{minipage}
    \vspace{0.3cm}
    
    \begin{minipage}[t]{\textwidth}
      \centering
      \captionof{table}{The experiments with different architectures for the global and personalized models. Green indicates MH-pFedHN, red indicates MH-pFedHNGD.}
      \label{table_ho_diff_global}
      \vspace{-0.2cm}
      \belowrulesep=0pt
      \aboverulesep=0pt
      \begin{small}
        \begin{sc}
          \scalebox{0.66}{
            \begin{tabular}{l|ccc|ccc}  
              \toprule
              & \multicolumn{6}{c}{CIFAR-100}   \\ 
              \cmidrule(lr){2-7} 
              \textnormal{Data\_Distribution}
              & \multicolumn{3}{c|}{\textnormal{non-IID\_1}} 
              & \multicolumn{3}{c}{\textnormal{non-IID\_2}} \\
              \cmidrule(lr){2-7} 
              \# \textnormal{Clients} & 50 & 100 & 200 & 50 & 100 & 200  \\ 
              \midrule  
              \rowcolor{LightGreen}
              \textnormal{VGGNet}  & 60.04 & 50.67 & 39.35 & 77.72 & 77.71 & 76.94 \\
              \rowcolor{LightGreen}
              \textnormal{ResNet}  & 62.03 & 55.38 & 42.09 & 75.57 & 72.50 & 65.50 \\
              \rowcolor{LightGreen}
              \textnormal{MLP}  & 52.14 & 49.09 & 43.48 & 69.97 & 73.13 & 74.32 \\
              \rowcolor{LightGreen}
              \textnormal{VGGNet/ResNet/MLP}  & 56.16 & 48.94 & 42.11 & 73.73 & 75.32 & 75.01 \\
              \rowcolor{LightRed}
              \textnormal{VGGNet}  & 64.75 & 61.03 & 44.65 & 79.82 & 79.93 & 78.07 \\
              \rowcolor{LightRed}
              \textnormal{ResNet}  & 64.47 & 55.40 & 49.43 & 75.66 & 74.59 & 73.67 \\
              \rowcolor{LightRed}
              \textnormal{MLP}  & 52.20 & 49.17 & 45.37 & 69.98 & 73.58 & 74.75 \\
              \rowcolor{LightRed}
              \textnormal{VGGNet/ResNet/MLP}  & 59.11 & 50.05 & 43.72 & 74.66 & 76.68 & 76.25 \\
              \bottomrule
            \end{tabular}
          }
        \end{sc}
      \end{small}
      \vspace{-0.5cm}
    \end{minipage}
  \end{minipage}
\end{figure}

\textbf{Generalize to new architectures.}
All the other baselines could not handle this most challenging case. We use new ResNet and SqueezeNet1\_1 for the newly added clients, while the training models are all different architectures. In this case, only the feature extractor of the hypernetwork is frozen; the new head will be added and trained. Figure~\ref{more_new} shows that no significant difference between MH-pFedHN and MH-pFedHNGD. The assistance provided by the global model is limited, and the learning during the generalization process primarily relies on the data distribution itself. Still, our methods outperform baselines.

\subsection{Ablation Study}

First, we investigate the head components of MH-pFedHN. We have two configurations: (1) MH-pFedHN does not have any heads; the hypernetwork consists of only a feature extractor; (2) MH-pFedHN has one head with an output dimension of $N$. Table~\ref{table_a_MH-pFedHN} indicates that without the head, the performance degrades significantly, demonstrating the importance of our head architecture in generating effective parameters. When using only a single head, most settings encounter out-of-memory issues, and the rest two cases show limited performance, which highlights that our carefully designed shared head can greatly reduce computation overhead and hold strong practical potential.

Next, we examine the impact of the global model. We use MH-pFedHNG to denote MH-pFedHNGD without knowledge distillation. Table~\ref{table_a} shows that in the vast majority of scenarios, MH-pFedHNGD outperforms MH-pFedHNG, which in turn outperforms MH-pFedHN. This demonstrates that the global model alone can enhance the generalization ability of the hypernetwork, while knowledge distillation with the global model further improves model training across heterogeneous clients. The observation that the results in homogeneous settings are better than those in heterogeneous ones is also consistent with our other experiments.

\subsection{Experiments with Different Architectures for Global Model and Personalized Clients}
\label{per_glo_diff}

In the previous experiments with heterogeneous models, some clients and the global model also used the same  LeNet-style architecture. Therefore, we further investigate the scenario where the architectures of the global model and the personalized client models are completely different. To this end, we set up experiments where the personalized client models are all VGGNet, MLP, and ResNet, while the global model is the LeNet-style model. We also consider the case where these three types of models are evenly distributed among the clients, with the global model still being the LeNet-style model. As shown in Table~\ref{table_ho_diff_global}, it can be observed that for MH-pFedHNGD, even when the architectures of the global and client models are completely different, its performance still surpasses that of MH-pFedHN, which is consistent with our previous findings. Under such a challenging scenario, both models achieve promising results, demonstrating the strong robustness and generalization ability of our method.

\subsection{Supplementary Experiments}
More experiments with the CIFAR-10 dataset, different client participation ratios, different architectures of global model, shared heads for heterogeneous models with similar parameter sizes, overhead, communication efficiency, iDLG Attacks, extremely personalized setting, and resource constraint setting are in Appendix~\ref{append:expexp}. Hyperparameter choice experiments could be found in Appendix~\ref{append:designlabels}.

\section{Conclusion}
In this work, we propose a novel data-free approach for model-heterogeneous personalized federated learning, termed MH-pFedHN. This method employs a hypernetwork located on the server to generate the model parameters required by each client, allowing clients to customize their network architectures without exposing them to the server. Furthermore, we present an improved MH-pFedHN with minimal effort, denoted as MH-pFedHNGD, which significantly enhances the learning and generalization abilities of the hypernetwork, improves the learning effectiveness of personalized models for clients, and reduces the risk of overfitting. 
We validate the effectiveness of our approaches through extensive experiments conducted in various settings. 
We note that switching from homogeneous models to heterogeneous models within the same settings can lead to a decline in accuracy, which represents a challenge that we aim to address in future research.


\begin{thebibliography}{10}

\bibitem{zhou2024adaptive}
Xiaokang Zhou, Wei Liang, Akira Kawai, Kaoru Fueda, Jinhua She, I~Kevin, and Kai Wang.
\newblock Adaptive segmentation enhanced asynchronous federated learning for sustainable intelligent transportation systems.
\newblock {\em IEEE Transactions on Intelligent Transportation Systems}, 2024.

\bibitem{wang2022ai}
Xiaoding Wang, Wenxin Liu, Hui Lin, Jia Hu, Kuljeet Kaur, and M~Shamim Hossain.
\newblock Ai-empowered trajectory anomaly detection for intelligent transportation systems: A hierarchical federated learning approach.
\newblock {\em IEEE Transactions on Intelligent Transportation Systems}, 24(4):4631--4640, 2022.

\bibitem{murmu2024reliable}
Anita Murmu, Piyush Kumar, Nageswara~Rao Moparthi, Suyel Namasudra, and Pascal Lorenz.
\newblock Reliable federated learning with gan model for robust and resilient future healthcare system.
\newblock {\em IEEE Transactions on Network and Service Management}, 2024.

\bibitem{ouyang2023harmony}
Xiaomin Ouyang, Zhiyuan Xie, Heming Fu, Sitong Cheng, Li~Pan, Neiwen Ling, Guoliang Xing, Jiayu Zhou, and Jianwei Huang.
\newblock Harmony: Heterogeneous multi-modal federated learning through disentangled model training.
\newblock In {\em Proceedings of the 21st Annual International Conference on Mobile Systems, Applications and Services}, pages 530--543, 2023.

\bibitem{nguyen2022federated}
Dinh~C Nguyen, Quoc-Viet Pham, Pubudu~N Pathirana, Ming Ding, Aruna Seneviratne, Zihuai Lin, Octavia Dobre, and Won-Joo Hwang.
\newblock Federated learning for smart healthcare: A survey.
\newblock {\em ACM Computing Surveys (Csur)}, 55(3):1--37, 2022.

\bibitem{feng2024robust}
Chenyuan Feng, Daquan Feng, Guanxin Huang, Zuozhu Liu, Zhenzhong Wang, and Xiang-Gen Xia.
\newblock Robust privacy-preserving recommendation systems driven by multimodal federated learning.
\newblock {\em IEEE Transactions on Neural Networks and Learning Systems}, 2024.

\bibitem{yuan2023federated}
Wei Yuan, Hongzhi Yin, Fangzhao Wu, Shijie Zhang, Tieke He, and Hao Wang.
\newblock Federated unlearning for on-device recommendation.
\newblock In {\em Proceedings of the sixteenth ACM international conference on web search and data mining}, pages 393--401, 2023.

\bibitem{guo2021prefer}
Yeting Guo, Fang Liu, Zhiping Cai, Hui Zeng, Li~Chen, Tongqing Zhou, and Nong Xiao.
\newblock Prefer: Point-of-interest recommendation with efficiency and privacy-preservation via federated edge learning.
\newblock {\em Proceedings of the ACM on Interactive, Mobile, Wearable and Ubiquitous Technologies}, 5(1):1--25, 2021.

\bibitem{10.5555/3294996.3295196}
Virginia Smith, Chao-Kai Chiang, Maziar Sanjabi, and Ameet Talwalkar.
\newblock Federated multi-task learning.
\newblock In {\em Proceedings of the 31st International Conference on Neural Information Processing Systems}, NIPS'17. Curran Associates Inc., 2017.

\bibitem{t2020personalized}
Canh T~Dinh, Nguyen Tran, and Josh Nguyen.
\newblock Personalized federated learning with moreau envelopes.
\newblock {\em Advances in neural information processing systems}, 33, 2020.

\bibitem{deng2020adaptive}
Yuyang Deng, Mohammad~Mahdi Kamani, and Mehrdad Mahdavi.
\newblock Adaptive personalized federated learning.
\newblock {\em arXiv preprint arXiv:2003.13461}, 2020.

\bibitem{McMahan2016CommunicationEfficientLO}
H.~B. McMahan, Eider Moore, Daniel Ramage, Seth Hampson, and Blaise~Ag{\"u}era y~Arcas.
\newblock Communication-efficient learning of deep networks from decentralized data.
\newblock In {\em International Conference on Artificial Intelligence and Statistics}, 2016.

\bibitem{chai2020tifl}
Zheng Chai, Ahsan Ali, Syed Zawad, Stacey Truex, Ali Anwar, Nathalie Baracaldo, Yi~Zhou, Heiko Ludwig, Feng Yan, and Yue Cheng.
\newblock Tifl: A tier-based federated learning system.
\newblock In {\em Proceedings of the 29th international symposium on high-performance parallel and distributed computing}, pages 125--136, 2020.

\bibitem{shin2024effective}
Yujin Shin, Kichang Lee, Sungmin Lee, You~Rim Choi, Hyung-Sin Kim, and JeongGil Ko.
\newblock Effective heterogeneous federated learning via efficient hypernetwork-based weight generation.
\newblock In {\em Proceedings of the 22nd ACM Conference on Embedded Networked Sensor Systems}, pages 112--125, 2024.

\bibitem{Caldas2018ExpandingTR}
Sebastian Caldas, Jakub Konecn{\'y}, H.~B. McMahan, and Ameet Talwalkar.
\newblock Expanding the reach of federated learning by reducing client resource requirements.
\newblock {\em ArXiv}, abs/1812.07210, 2018.

\bibitem{10.5555/3495724.3496904}
Chaoyang He, Murali Annavaram, and Salman Avestimehr.
\newblock Group knowledge transfer: federated learning of large cnns at the edge.
\newblock In {\em Proceedings of the 34th International Conference on Neural Information Processing Systems}, Red Hook, NY, USA, 2020. Curran Associates Inc.

\bibitem{Shah2021ModelCF}
Suhail~Mohmad Shah and Vincent K.~N. Lau.
\newblock Model compression for communication efficient federated learning.
\newblock {\em IEEE Transactions on Neural Networks and Learning Systems}, 34:5937--5951, 2021.

\bibitem{li2019fedmd}
Daliang Li and Junpu Wang.
\newblock Fedmd: Heterogenous federated learning via model distillation.
\newblock {\em arXiv preprint arXiv:1910.03581}, 2019.

\bibitem{zhu2021data}
Zhuangdi Zhu, Junyuan Hong, and Jiayu Zhou.
\newblock Data-free knowledge distillation for heterogeneous federated learning.
\newblock In {\em International conference on machine learning}, pages 12878--12889. PMLR, 2021.

\bibitem{wu2024exploring}
Zhiyuan Wu, Sheng Sun, Yuwei Wang, Min Liu, Quyang Pan, Junbo Zhang, Zeju Li, and Qingxiang Liu.
\newblock Exploring the distributed knowledge congruence in proxy-data-free federated distillation.
\newblock {\em ACM Transactions on Intelligent Systems and Technology}, 15(2):1--34, 2024.

\bibitem{Chen2023EfficientPF}
Daoyuan Chen, Liuyi Yao, Dawei Gao, Bolin Ding, and Yaliang Li.
\newblock Efficient personalized federated learning via sparse model-adaptation.
\newblock {\em ArXiv}, abs/2305.02776, 2023.

\bibitem{diao2021heterofl}
Enmao Diao, Jie Ding, and Vahid Tarokh.
\newblock Hetero{\{}fl{\}}: Computation and communication efficient federated learning for heterogeneous clients.
\newblock In {\em International Conference on Learning Representations}, 2021.

\bibitem{alam2022fedrolex}
Samiul Alam, Luyang Liu, Ming Yan, and Mi~Zhang.
\newblock Fedrolex: Model-heterogeneous federated learning with rolling sub-model extraction.
\newblock {\em Advances in neural information processing systems}, 35:29677--29690, 2022.

\bibitem{hong2022efficientsplitmixfederatedlearning}
Junyuan Hong, Haotao Wang, Zhangyang Wang, and Jiayu Zhou.
\newblock Efficient split-mix federated learning for on-demand and in-situ customization, 2022.

\bibitem{10.1145/3643832.3661880}
Leming Shen, Qiang Yang, Kaiyan Cui, Yuanqing Zheng, Xiao-Yong Wei, Jianwei Liu, and Jinsong Han.
\newblock Fedconv: A learning-on-model paradigm for heterogeneous federated clients.
\newblock In {\em Proceedings of the 22nd Annual International Conference on Mobile Systems, Applications and Services}, MOBISYS '24, page 398–411, New York, NY, USA, 2024. Association for Computing Machinery.

\bibitem{Zhang2023TowardsDK}
J.~Zhang, Song Guo, Jingcai Guo, Deze Zeng, Jingren Zhou, and Albert~Y. Zomaya.
\newblock Towards data-independent knowledge transfer in model-heterogeneous federated learning.
\newblock {\em IEEE Transactions on Computers}, 72:2888--2901, 2023.

\bibitem{wang2024dfrd}
Shuai Wang, Yexuan Fu, Xiang Li, Yunshi Lan, Ming Gao, et~al.
\newblock Dfrd: Data-free robustness distillation for heterogeneous federated learning.
\newblock {\em Advances in Neural Information Processing Systems}, 36, 2024.

\bibitem{liu2025adapter}
Shichong Liu, Haozhe Jin, Zhiwei Tang, Rui Zhai, Ke~Lu, Junyang Yu, and Chenxi Bai.
\newblock Adapter-guided knowledge transfer for heterogeneous federated learning.
\newblock {\em Journal of Systems Architecture}, page 103338, 2025.

\bibitem{10734591}
Sheng Guo, Hui Chen, Yang Liu, Chengyi Yang, Zengxiang Li, and Cheng~Hao Jin.
\newblock Heterogeneous federated learning framework for iiot based on selective knowledge distillation.
\newblock {\em IEEE Transactions on Industrial Informatics}, 21(2):1078--1089, 2025.

\bibitem{xu2023personalized}
Jian Xu, Xinyi Tong, and Shao-Lun Huang.
\newblock Personalized federated learning with feature alignment and classifier collaboration.
\newblock {\em arXiv preprint arXiv:2306.11867}, 2023.

\bibitem{pmlr-v139-collins21a}
Liam Collins, Hamed Hassani, Aryan Mokhtari, and Sanjay Shakkottai.
\newblock Exploiting shared representations for personalized federated learning.
\newblock In Marina Meila and Tong Zhang, editors, {\em Proceedings of the 38th International Conference on Machine Learning}, volume 139 of {\em Proceedings of Machine Learning Research}, pages 2089--2099. PMLR, 18--24 Jul 2021.

\bibitem{arivazhagan2019federated}
Manoj~Ghuhan Arivazhagan, Vinay Aggarwal, Aaditya~Kumar Singh, and Sunav Choudhary.
\newblock Federated learning with personalization layers.
\newblock {\em arXiv preprint arXiv:1912.00818}, 2019.

\bibitem{liang2020thinklocallyactglobally}
Paul~Pu Liang, Terrance Liu, Liu Ziyin, Nicholas~B. Allen, Randy~P. Auerbach, David Brent, Ruslan Salakhutdinov, and Louis-Philippe Morency.
\newblock Think locally, act globally: Federated learning with local and global representations, 2020.

\bibitem{10.1145/3545008.3545073}
Jaehee Jang, Heoneok Ha, Dahuin Jung, and Sungroh Yoon.
\newblock Fedclassavg: Local representation learning for personalized federated learning on heterogeneous neural networks.
\newblock In {\em Proceedings of the 51st International Conference on Parallel Processing}, ICPP '22, New York, NY, USA, 2023. Association for Computing Machinery.

\bibitem{10.1145/3581783.3611781}
Liping Yi, Gang Wang, Xiaoguang Liu, Zhuan Shi, and Han Yu.
\newblock Fedgh: Heterogeneous federated learning with generalized global header.
\newblock In {\em Proceedings of the 31st ACM International Conference on Multimedia}, MM '23, page 8686¨C8696, New York, NY, USA, 2023. Association for Computing Machinery.

\bibitem{yi2023pfedes}
Liping Yi, Han Yu, Gang Wang, and Xiaoguang Liu.
\newblock pfedes: Model heterogeneous personalized federated learning with feature extractor sharing.
\newblock {\em arXiv preprint arXiv:2311.06879}, 2023.

\bibitem{NEURIPS2021_6aed000a}
Samuel Horv\'{a}th, Stefanos Laskaridis, Mario Almeida, Ilias Leontiadis, Stylianos Venieris, and Nicholas Lane.
\newblock Fjord: Fair and accurate federated learning under heterogeneous targets with ordered dropout.
\newblock In M.~Ranzato, A.~Beygelzimer, Y.~Dauphin, P.S. Liang, and J.~Wortman Vaughan, editors, {\em Advances in Neural Information Processing Systems}, volume~34, pages 12876--12889. Curran Associates, Inc., 2021.

\bibitem{Ilhan_2023_CVPR}
Fatih Ilhan, Gong Su, and Ling Liu.
\newblock Scalefl: Resource-adaptive federated learning with heterogeneous clients.
\newblock In {\em Proceedings of the IEEE/CVF Conference on Computer Vision and Pattern Recognition (CVPR)}, pages 24532--24541, June 2023.

\bibitem{lee2024recurrentearlyexitsfederated}
Royson Lee, Javier Fernandez-Marques, Shell~Xu Hu, Da~Li, Stefanos Laskaridis, Łukasz Dudziak, Timothy Hospedales, Ferenc Huszár, and Nicholas~D. Lane.
\newblock Recurrent early exits for federated learning with heterogeneous clients, 2024.

\bibitem{Gong_2021_ICCV}
Xuan Gong, Abhishek Sharma, Srikrishna Karanam, Ziyan Wu, Terrence Chen, David Doermann, and Arun Innanje.
\newblock Ensemble attention distillation for privacy-preserving federated learning.
\newblock In {\em Proceedings of the IEEE/CVF International Conference on Computer Vision (ICCV)}, pages 15076--15086, October 2021.

\bibitem{itahara2021distillation}
Sohei Itahara, Takayuki Nishio, Yusuke Koda, Masahiro Morikura, and Koji Yamamoto.
\newblock Distillation-based semi-supervised federated learning for communication-efficient collaborative training with non-iid private data.
\newblock {\em IEEE Transactions on Mobile Computing}, 22(1):191--205, 2021.

\bibitem{Wang_Yan_Wang_Wang_Shu_Cheng_Chen_2025}
Zichen Wang, Feng Yan, Tianyi Wang, Cong Wang, Yuanchao Shu, Peng Cheng, and Jiming Chen.
\newblock Fed-dfa: Federated distillation for heterogeneous model fusion through the adversarial lens.
\newblock {\em Proceedings of the AAAI Conference on Artificial Intelligence}, 39(20):21429--21437, Apr. 2025.

\bibitem{ha2017hypernetworks}
David Ha, Andrew~M. Dai, and Quoc~V. Le.
\newblock Hypernetworks.
\newblock In {\em International Conference on Learning Representations}, 2017.

\bibitem{hinton2015distilling}
Geoffrey Hinton.
\newblock Distilling the knowledge in a neural network.
\newblock {\em arXiv preprint arXiv:1503.02531}, 2015.

\bibitem{jeong2018communication}
Eunjeong Jeong, Seungeun Oh, Hyesung Kim, Jihong Park, Mehdi Bennis, and Seong-Lyun Kim.
\newblock Communication-efficient on-device machine learning: Federated distillation and augmentation under non-iid private data.
\newblock {\em arXiv preprint arXiv:1811.11479}, 2018.

\bibitem{von2020continual}
Johannes von Oswald, Christian Henning, Benjamin~F Grewe, and Jo{\~a}o Sacramento.
\newblock Continual learning with hypernetworks.
\newblock In {\em 8th International Conference on Learning Representations (ICLR 2020)(virtual)}. International Conference on Learning Representations, 2020.

\bibitem{suarez2017language}
Joseph Suarez.
\newblock Language modeling with recurrent highway hypernetworks.
\newblock {\em Advances in neural information processing systems}, 30, 2017.

\bibitem{nirkin2021hyperseg}
Yuval Nirkin, Lior Wolf, and Tal Hassner.
\newblock Hyperseg: Patch-wise hypernetwork for real-time semantic segmentation.
\newblock In {\em Proceedings of the IEEE/CVF conference on computer vision and pattern recognition}, 2021.

\bibitem{beck2024recurrent}
Jacob Beck, Risto Vuorio, Zheng Xiong, and Shimon Whiteson.
\newblock Recurrent hypernetworks are surprisingly strong in meta-rl.
\newblock {\em Advances in Neural Information Processing Systems}, 36, 2024.

\bibitem{ruiz2024hyperdreambooth}
Nataniel Ruiz, Yuanzhen Li, Varun Jampani, Wei Wei, Tingbo Hou, Yael Pritch, Neal Wadhwa, Michael Rubinstein, and Kfir Aberman.
\newblock Hyperdreambooth: Hypernetworks for fast personalization of text-to-image models.
\newblock In {\em Proceedings of the IEEE/CVF Conference on Computer Vision and Pattern Recognition}, pages 6527--6536, 2024.

\bibitem{shamsian2021personalized}
Aviv Shamsian, Aviv Navon, Ethan Fetaya, and Gal Chechik.
\newblock Personalized federated learning using hypernetworks.
\newblock In {\em International Conference on Machine Learning}, pages 9489--9502. PMLR, 2021.

\bibitem{zhu2023layer}
Suxia Zhu, Tianyu Liu, and Guanglu Sun.
\newblock Layer-wise personalized federated learning with hypernetwork.
\newblock {\em Neural Processing Letters}, 55(9):12273--12287, 2023.

\bibitem{scott2024pefll}
Jonathan~A Scott, Hossein Zakerinia, and Christoph Lampert.
\newblock Pefll: Personalized federated learning by learning to learn.
\newblock In {\em 12th International Conference on Learning Representations}, 2024.

\bibitem{ma2022layer}
Xiaosong Ma, Jie Zhang, Song Guo, and Wenchao Xu.
\newblock Layer-wised model aggregation for personalized federated learning.
\newblock In {\em Proceedings of the IEEE/CVF conference on computer vision and pattern recognition}, pages 10092--10101, 2022.

\bibitem{sendera2023hypershot}
Marcin Sendera, Marcin Przewiezlikowski, Konrad Karanowski, Maciej Zieba, Jacek Tabor, and Przemyslaw Spurek.
\newblock Hypershot: Few-shot learning by kernel hypernetworks.
\newblock In {\em 2023 IEEE/CVF Winter Conference on Applications of Computer Vision (WACV)}, pages 2468--2477. IEEE Computer Society, 2023.

\bibitem{Litany2022FederatedLW}
Or~Litany, Haggai Maron, David Acuna, Jan Kautz, Gal Chechik, and Sanja Fidler.
\newblock Federated learning with heterogeneous architectures using graph hypernetworks.
\newblock {\em ArXiv}, abs/2201.08459, 2022.

\bibitem{li2021fedbn}
Xiaoxiao Li, Meirui JIANG, Xiaofei Zhang, Michael Kamp, and Qi~Dou.
\newblock Fed{BN}: Federated learning on non-{IID} features via local batch normalization.
\newblock In {\em International Conference on Learning Representations}, 2021.

\bibitem{10.5555/3698900.3699280}
Guangsheng Zhang, Bo~Liu, Huan Tian, Tianqing Zhu, Ming Ding, and Wanlei Zhou.
\newblock How does a deep learning model architecture impact its privacy? a comprehensive study of privacy attacks on cnns and transformers.
\newblock In {\em Proceedings of the 33rd USENIX Conference on Security Symposium}, SEC '24, USA, 2024. USENIX Association.

\bibitem{kim2023depthfl}
Minjae Kim, Sangyoon Yu, Suhyun Kim, and Soo-Mook Moon.
\newblock Depthfl: Depthwise federated learning for heterogeneous clients.
\newblock In {\em The Eleventh International Conference on Learning Representations}, 2023.

\bibitem{sattler2020communication}
Felix Sattler, Arturo Marban, Roman Rischke, and Wojciech Samek.
\newblock Communication-efficient federated distillation.
\newblock {\em arXiv preprint arXiv:2012.00632}, 2020.

\bibitem{DBLP:journals/corr/abs-2108-13323}
Chuhan Wu, Fangzhao Wu, Ruixuan Liu, Lingjuan Lyu, Yongfeng Huang, and Xing Xie.
\newblock Fedkd: Communication efficient federated learning via knowledge distillation.
\newblock {\em CoRR}, abs/2108.13323, 2021.

\bibitem{lin2020ensemble}
Tao Lin, Lingjing Kong, Sebastian~U Stich, and Martin Jaggi.
\newblock Ensemble distillation for robust model fusion in federated learning.
\newblock {\em Advances in neural information processing systems}, 33:2351--2363, 2020.

\bibitem{zhang2021parameterized}
Jie Zhang, Song Guo, Xiaosong Ma, Haozhao Wang, Wenchao Xu, and Feijie Wu.
\newblock Parameterized knowledge transfer for personalized federated learning.
\newblock {\em Advances in Neural Information Processing Systems}, 34:10092--10104, 2021.

\bibitem{zhang2022dense}
Jie Zhang, Chen Chen, Bo~Li, Lingjuan Lyu, Shuang Wu, Shouhong Ding, Chunhua Shen, and Chao Wu.
\newblock Dense: Data-free one-shot federated learning.
\newblock {\em Advances in Neural Information Processing Systems}, 35:21414--21428, 2022.

\bibitem{dai2024enhancing}
Rong Dai, Yonggang Zhang, Ang Li, Tongliang Liu, Xun Yang, and Bo~Han.
\newblock Enhancing one-shot federated learning through data and ensemble co-boosting.
\newblock {\em arXiv preprint arXiv:2402.15070}, 2024.

\bibitem{zhang2022fedzkt}
Lan Zhang, Dapeng Wu, and Xiaoyong Yuan.
\newblock Fedzkt: Zero-shot knowledge transfer towards resource-constrained federated learning with heterogeneous on-device models.
\newblock In {\em 2022 IEEE 42nd International Conference on Distributed Computing Systems (ICDCS)}, pages 928--938. IEEE, 2022.

\bibitem{goodfellow2020generative}
Ian Goodfellow, Jean Pouget-Abadie, Mehdi Mirza, Bing Xu, David Warde-Farley, Sherjil Ozair, Aaron Courville, and Yoshua Bengio.
\newblock Generative adversarial networks.
\newblock {\em Communications of the ACM}, 63(11):139--144, 2020.

\bibitem{10.5555/3666122.3666906}
Kangyang Luo, Shuai Wang, Yexuan Fu, Xiang Li, Yunshi Lan, and Ming Gao.
\newblock Dfrd: data-free robustness distillation for heterogeneous federated learning.
\newblock In {\em Proceedings of the 37th International Conference on Neural Information Processing Systems}, NIPS '23, Red Hook, NY, USA, 2024. Curran Associates Inc.

\bibitem{Csiszr1975IDivergenceGO}
Imre Csisz{\'a}r.
\newblock \$i\$-divergence geometry of probability distributions and minimization problems.
\newblock {\em Annals of Probability}, 3:146--158, 1975.

\bibitem{CohenATS17}
Gregory Cohen, Saeed Afshar, Jonathan Tapson, and Andr{\'{e}} van Schaik.
\newblock {EMNIST:} extending {MNIST} to handwritten letters.
\newblock In {\em 2017 International Joint Conference on Neural Networks, {IJCNN} 2017, Anchorage, AK, USA, May 14-19, 2017}, pages 2921--2926. {IEEE}, 2017.

\bibitem{2009Learning}
A.~Krizhevsky and G.~Hinton.
\newblock Learning multiple layers of features from tiny images.
\newblock {\em Handbook of Systemic Autoimmune Diseases}, 1(4), 2009.

\bibitem{le2015tiny}
Yann Le and Xuan Yang.
\newblock Tiny imagenet visual recognition challenge.
\newblock {\em CS 231N}, 7(7):3, 2015.

\bibitem{li2022federated}
Qinbin Li, Yiqun Diao, Quan Chen, and Bingsheng He.
\newblock Federated learning on non-iid data silos: An experimental study.
\newblock In {\em IEEE International Conference on Data Engineering}, 2022.

\bibitem{liu2024fedlpa}
Xiang Liu, Liangxi Liu, Feiyang Ye, Yunheng Shen, Xia Li, Linshan Jiang, and Jialin Li.
\newblock Fedlpa: One-shot federated learning with layer-wise posterior aggregation.
\newblock {\em Advances in Neural Information Processing Systems}, 37:81510--81548, 2024.

\bibitem{lecun1998gradient}
Yann LeCun, L{\'e}on Bottou, Yoshua Bengio, and Patrick Haffner.
\newblock Gradient-based learning applied to document recognition.
\newblock {\em Proceedings of the IEEE}, 86(11):2278--2324, 1998.

\bibitem{simonyan2014very}
Karen Simonyan and Andrew Zisserman.
\newblock Very deep convolutional networks for large-scale image recognition.
\newblock In {\em International Conference on Learning Representations}, 2015.

\bibitem{He_2016_CVPR}
Kaiming He, Xiangyu Zhang, Shaoqing Ren, and Jian Sun.
\newblock Deep residual learning for image recognition.
\newblock In {\em Proceedings of the IEEE Conference on Computer Vision and Pattern Recognition (CVPR)}, June 2016.

\bibitem{zagoruyko2016wide}
Sergey Zagoruyko.
\newblock Wide residual networks.
\newblock {\em arXiv preprint arXiv:1605.07146}, 2016.

\bibitem{mcmahan2017communication}
Brendan McMahan, Eider Moore, Daniel Ramage, Seth Hampson, and Blaise~Aguera y~Arcas.
\newblock Communication-efficient learning of deep networks from decentralized data.
\newblock In {\em Artificial intelligence and statistics}, pages 1273--1282. PMLR, 2017.

\bibitem{Kingma2014AdamAM}
Diederik~P. Kingma and Jimmy Ba.
\newblock Adam: A method for stochastic optimization.
\newblock {\em CoRR}, abs/1412.6980, 2014.

\bibitem{iandola2016squeezenet}
Forrest~N Iandola.
\newblock Squeezenet: Alexnet-level accuracy with 50x fewer parameters and< 0.5 mb model size.
\newblock {\em arXiv preprint arXiv:1602.07360}, 2016.

\bibitem{Zhao2020iDLGID}
Bo~Zhao, Konda~Reddy Mopuri, and Hakan Bilen.
\newblock idlg: Improved deep leakage from gradients.
\newblock {\em ArXiv}, abs/2001.02610, 2020.

\end{thebibliography}

\newpage
\appendix

\addcontentsline{toc}{section}{Appendix} %
\part{Appendix} %
\parttoc %

\begin{figure}[H] 
\centering
\begin{minipage}[t]{0.48\textwidth} 
\begin{algorithm}[H]
  \caption{Model-Heterogeneous Personalized Federated Hypernetwork}
   \label{alg1}  
            \begin{algorithmic}  
                \Require{$R$ - number of rounds, $\alpha$ - HN learning rate, $\eta$ - client learning rate, $E$ - client local epoch, $\{K_1,\dots,K_n\}$ - number of clients parameter, $\{D_1,\dots,D_n\}$ - datasets}  
                \Ensure{trained model parameters $\{\bm{\theta}_1,\dots,\bm{\theta}_n\}$}  
                \Procedure{Server executes}{}  
                \For{$r=1$ {\bfseries to} $R$}  
                \For{each client $i$ \textbf{in parallel}}  
                \State $\bm{\theta}_i = h(\bm{v}_i;\varphi)_{[1:K_i]}$  
                \State $\Delta\bm{\theta}_{i} \leftarrow \text{ClientUpdate}(\bm{\theta}_i)$  
                \State $\varphi = \varphi - \alpha\nabla_{\varphi}\bm{\theta}^{T}_{i}\Delta\bm{\theta}_i$  
                \State $\bm{v}_i = \bm{v}_i - \alpha\nabla_{\bm{v}_i}\varphi^{T}\nabla_{\varphi}\bm{\theta}^{T}_{i}\Delta\bm{\theta}_i$  
                \EndFor	  
                \EndFor	  
                \EndProcedure  
				  
                \Function{ClientUpdate}{$\bm{\theta}_i$}  
                \State  $\widetilde{\bm{\theta}_i} = \bm{\theta}_i$  	  
                \For{$e=1$ {\bfseries to} $E$}  
                \State sample batch $B \subset D_i$   
                \State $\widetilde{\bm{\theta}_i} = \widetilde{\bm{\theta}_i} - \eta\nabla_{\widetilde{\bm{\theta}_i}}L(\widetilde{\bm{\theta}_i},B)$  
                \EndFor				  
                \State $\Delta\bm{\theta}_{i} = \widetilde{\bm{\theta_i}} - \bm{\theta}_{i}$	  
                \State \Return{$\Delta\bm{\theta}_{i}$}  
                \EndFunction  
  \end{algorithmic}
\end{algorithm}
\end{minipage}
\hfill
\begin{minipage}[t]{0.48\textwidth} 
\begin{algorithm}[H]
  \caption{Model-Heterogeneous Personalized Federated Hypernetwork with Global Distillation}  
            \label{alg2}  
            \begin{algorithmic}  
                \Require{$R$ - number of rounds, $\alpha$ - HN learning rate, $\eta$ - client learning rate, $E$ - client local epoch, $\{K_1,\dots,K_n\}$ - number of clients parameter, $\{D_1,\dots,D_n\}$ - datasets, $\lambda$ - balancing factor of distillation loss}  
                \Ensure{trained model parameters $\{\bm{\theta}_1,\dots,\bm{\theta}_n\}$}  
                \Procedure{Server executes}{}  
                \For{$r=1$ {\bfseries to} $R$}  
                \State $\bm{w}_g = h(\bm{v}_g;\varphi)_{[1:K_g]}$   
                \For{each client $i$ \textbf{in parallel}}  
                \State $ \Delta \bm{w}_{g,i} \leftarrow \text{ClientUpdate1}(\bm{w}_g)$  
                \EndFor	  
                \State $\varphi = \varphi - \alpha \sum_{i=1}^n \frac{m_i}{M} \nabla_{\varphi}\bm{w}^{T}_{g}\Delta \bm{w}_{g,i}$  
                \State $\bm{v}_g = \bm{v}_g - \alpha \sum_{i=1}^n \frac{m_i}{M} \nabla_{\bm{v}_g}\varphi^{T}\nabla_{\varphi}\bm{w}^{T}_{g}\Delta \bm{w}_{g,i}$  

                \For{each client $i$ \textbf{in parallel}}  
                \State $\bm{\theta}_i = h(\bm{v}_i;\varphi)_{[1:K_i]}$  
                \State $\Delta\bm{\theta}_{i} \leftarrow \text{ClientUpdate2}(\bm{\theta}_i)$  
                \State $\varphi = \varphi - \alpha\nabla_{\varphi}\bm{\theta}^{T}_{i}\Delta\bm{\theta}_i$  
                \State $\bm{v}_i = \bm{v}_i - \alpha\nabla_{\bm{v}_i}\varphi^{T}\nabla_{\varphi}\bm{\theta}^{T}_{i}\Delta\bm{\theta}_i$  
                \EndFor	  
                \EndFor
                \EndProcedure  
                
                \Function{ClientUpdate1}{$\bm{w}_g$}  
                \State $\bm{w}_{g,i} = \bm{w}_g$  
                \For{$e=1$ {\bfseries to} $E$}  
                \State sample batch $B \subset D_i$   
                \State $\bm{w}_{g,i} = \bm{w}_{g,i} - \eta\nabla_{\bm{w}_{g,i}}L(\bm{w}_{g,i},B)$  
                \EndFor	  
                \State $\Delta \bm{w}_{g,i} = \bm{w}_{g,i} - \bm{w}_{g}$	  
                \State \Return{$\Delta \bm{w}_{g,i}$}  
                \EndFunction  
				  
                \Function{ClientUpdate2}{$\bm{\theta}_i$}  
                \State $\widetilde{\bm{\theta}_i} = \bm{\theta}_i$ 
                \For{$e=1$ {\bfseries to} $E$}  
                \State sample batch $B \subset D_i$   
                \State $\widetilde{\bm{\theta}_i} = \widetilde{\bm{\theta}_i} - \eta \nabla_{\widetilde{\bm{\theta}_i}} \left(\lambda L(\widetilde{\bm{\theta}_i}, B) + (1-\lambda)L_{KL}(\widetilde{\bm{\theta}_i}, \bm{w}_g, B) \right)$  
                \EndFor				  
                \State $\Delta\bm{\theta}_{i} = \widetilde{\bm{\theta}_i} - \bm{\theta}_{i}$  
                \State \Return{$\Delta\bm{\theta}_{i}$}  
                \EndFunction  
  \end{algorithmic}
\end{algorithm}
\end{minipage}
\end{figure}

\section{Algorithms}
\label{Algorithm_and_Analysis} 
In this section, we will outline the algorithms related to MH-pFedHN and MH-pFedHNGD (see Algorithms \ref{alg1} and \ref{alg2}). 

\section{Additional Experiment Settings}
\label{append:expsetting}

\subsection{Dataset}
We used four widely adopted datasets to evaluate our proposed methods: CIFAR-10~\cite{2009Learning}, CIFAR-100~\cite{2009Learning}, Tiny-ImageNet~\cite{le2015tiny}, and EMNIST~\cite{CohenATS17}. 
\begin{itemize}
    \item CIFAR-10. CIFAR-10 consists of 60,000 color images with $32 \times 32$ pixels, evenly distributed across 10 classes.
    \item CIFAR-100. CIFAR-100 dataset consists of 60,000 $32 \times 32$ color images, evenly distributed across 100 classes.
    \item Tiny-ImageNet. Tiny-ImageNet dataset contains 110,000 images at $64\times64$ pixels across 200 classes and is a more challenging dataset with 500 training and 50 validation images per class.
    \item EMNIST (ByClass). The EMNIST dataset is an extension of the MNIST dataset, providing a more extensive set of handwritten character images. In our experiments, we use the byclass subset, which contains 814,255 images of size $28 \times 28$ pixels across 62 classes (10 digits (0-9) and 52 letters (uppercase A-Z and lowercase a-z)).
\end{itemize}

\subsection{Implementation} 
\label{append:implementation}
We conduct simulations of all clients and the server on a workstation equipped with an RTX 4090 GPU, a 2.6-GHz Intel(R) Xeon(R) W7-2475X CPU, and 125 GiB of RAM. The implementation of all methods is done using PyTorch. 

\section{Additional Experiments}
\label{append:expexp}

\subsection{Additional Experiment with CIFAR-100/Tiny-ImageNet}
We provide additional experiments over the CIFAR-100/Tiny-ImageNet datasets. Here, we compare MH-pFedHN and MH-pFedHNGD to the baselines on a small-scale setup of 10 clients. The results are presented in Tables~\ref{table_10client_merged}. We show significant improvement using MH-pFedHN and MH-pFedHNGD on small-scale experiments in addition to the results presented in the main text.

\begin{table}[!t]
\caption{Comparison under Homogeneous and Heterogeneous Model Settings (10 Clients)}
\label{table_10client_merged}
\centering
\setlength{\tabcolsep}{4pt}
\scriptsize
\begin{sc}
\begin{tabular}{lcccccccc}
\toprule
& \multicolumn{4}{c}{\textnormal{\textbf{Homogeneous Model}}} & \multicolumn{4}{c}{\textnormal{\textbf{Heterogeneous Model}}} \\
\cmidrule(lr){2-5} 
\cmidrule(lr){6-9}
\multirow{2}{*}{\textnormal{Method}} & \multicolumn{2}{c}{CIFAR-100} & \multicolumn{2}{c}{\textnormal{Tiny-ImageNet}} & \multicolumn{2}{c}{CIFAR-100} & \multicolumn{2}{c}{\textnormal{Tiny-ImageNet}} \\
\cmidrule(lr){2-3} 
\cmidrule(lr){4-5} 
\cmidrule(lr){6-7} 
\cmidrule(lr){8-9} 
& \textnormal{non-IID\_1} & \textnormal{non-IID\_2} & \textnormal{non-IID\_1} & \textnormal{non-IID\_2} & \textnormal{non-IID\_1} & \textnormal{non-IID\_2} & \textnormal{non-IID\_1} & \textnormal{non-IID\_2} \\
\midrule
\textnormal{Local} & 64.63 & 61.53 & 39.94 & 38.50 & 65.59 & 63.33 & 42.59 & 41.89 \\
\textnormal{FedAvg~\cite{mcmahan2017communication}} & 29.70 & 23.83 & 9.99 & 7.05 & 27.06 & 19.12 & 9.59 & 7.48 \\
\midrule
\textnormal{pFedHN~\cite{shamsian2021personalized}} & 63.68 & 61.91 & 41.05 & 38.73 & 60.57 & 60.76 & 39.44 & 40.24 \\
\textnormal{pFedLA~\cite{ma2022layer}} & 65.56 & 56.79 & 43.53 & 34.30 & 56.53 & 55.13 & 35.57 & 33.61 \\
\textnormal{FedGH~\cite{10.1145/3581783.3611781}} & 65.84 & 63.34 & 41.34 & 39.43 & 63.04 & 61.65 & 36.48 & 31.84 \\
\textnormal{pFedLHN~\cite{zhu2023layer}} & 66.46 & 64.79 & 42.78 & 39.78 & 66.76 & 65.35 & 43.08 & 43.16 \\
\textnormal{PeFLL~\cite{scott2024pefll}} & 52.82 & 48.31 & 37.59 & 33.48 & 61.14 & 58.57 & 41.20 & 41.38 \\
\textnormal{FedAKT~\cite{liu2025adapter}} & \textbf{69.17} & 63.27 & 43.61 & 39.82 & 66.61 & 65.84 & 43.15 & 42.36 \\
\midrule
\textnormal{MH-pFedHN (ours)} & 68.12 & \textbf{64.81} & \textbf{44.38} & \textbf{40.59} & \textbf{67.02} & \textbf{66.46} & \textbf{44.80} & \textbf{45.14} \\
\textnormal{MH-pFedHNGD (ours)} & 68.39 & \textbf{64.98} & \textbf{44.49} & \textbf{41.62} & \textbf{67.10} & \textbf{67.41} & \textbf{45.51} & 44.28 \\
\bottomrule
\end{tabular}
\end{sc}
\end{table}

\subsection{Experiments with CIFAR-10}
\label{append:cifar10}
We provide additional experiments over the CIFAR-10 dataset. Table~\ref{merged_table_cifar10} shows that in both non-IID settings, MH-pFedHN and MH-pFedHNGD achieve superior accuracy compared to baseline methods, highlighting their effectiveness in handling data heterogeneity.

\begin{table}[!t]
\caption{Test accuracy (\%) over 10, 50, and 100 clients on CIFAR-10 under Homogeneous and Heterogeneous models with Heterogeneous data.}
\label{merged_table_cifar10}
\centering
\scriptsize
\setlength{\tabcolsep}{4pt}
\begin{sc}
\begin{tabular}{lcccccccccccr}
\toprule
& \multicolumn{6}{c}{\textnormal{Homogeneous Model}} & \multicolumn{6}{c}{\textnormal{Heterogeneous Model}} \\
\cmidrule(lr){2-7} \cmidrule(lr){8-13} 
\multirow{2}{*}{\textnormal{Method}} & \multicolumn{3}{c}{\textnormal{non-IID\_1}} & \multicolumn{3}{c}{\textnormal{non-IID\_2}} & \multicolumn{3}{c}{\textnormal{non-IID\_1}} & \multicolumn{3}{c}{\textnormal{non-IID\_2}} \\
\cmidrule(lr){2-4} \cmidrule(lr){5-7} \cmidrule(lr){8-10} \cmidrule(lr){11-13} 
& 10 & 50 & 100 & 10 & 50 & 100 & 10 & 50 & 100 & 10 & 50 & 100 \\
\midrule
\textnormal{Local} & 92.71 & 89.38 & 87.87 & 83.04 & 82.77 & 79.53 & 93.23 & 84.98 & 80.26 & 81.97 & 80.58 & 75.09 \\
\textnormal{FedAvg} & 50.15 & 51.39 & 58.28 & 64.23 & 64.77 & 61.65 & 54.59 & 65.51 & 57.28 & 73.61 & 56.75 & 51.65 \\
\midrule
\textnormal{pFedHN} & 92.60 & 91.66 & 90.87 & 82.67 & 85.49 & 81.78 & 92.83 & 88.11 & 87.00 & 81.12 & 80.96 & 77.75 \\
\textnormal{pFedLA} & 85.12 & 83.86 & 81.71 & 77.49 & 84.06 & 77.14 & 93.26 & 73.27 & 65.38 & 77.67 & 81.00 & 75.55 \\
\textnormal{FedGH} & 92.90 & 88.24 & 85.39 & 84.19 & 73.55 & 72.82 & 92.56 & 86.04 & 83.36 & 83.01 & 70.39 & 70.26 \\
\textnormal{pFedLHN} & 94.16 & 91.09 & 90.71 & 82.49 & 86.10 & 85.48 & 93.67 & 89.33 & 87.26 & 83.79 & 84.21 & 79.82 \\
\textnormal{PeFLL} & 89.21 & 86.62 & 86.08 & 75.59 & 78.25 & 78.76 & 92.21 & 89.09 & 88.20 & 80.85 & 82.49 & 80.65 \\
\textnormal{FedAKT} & 93.46 & 91.14 & 90.31 & 83.88 & 85.96 & 82.58 & 93.58 & 88.59 & 85.62 & 82.82 & 83.05 & 78.69 \\
\midrule
\textnormal{MH-pFedHN (ours)} & \textbf{94.24} & \textbf{91.69} & \textbf{91.23} & 83.72 & \textbf{87.13} & 85.79 & \textbf{93.98} & \textbf{89.74} & \textbf{88.53} & \textbf{83.87} & 84.11 & 80.53 \\
\textnormal{MH-pFedHNGD (ours)} & 93.56 & \textbf{91.75} & \textbf{92.12} & 82.92 & \textbf{87.50} & \textbf{86.25} & 93.81 & \textbf{90.26} & \textbf{88.73} & 83.83 & \textbf{85.52} & \textbf{80.88} \\
\bottomrule
\end{tabular}
\end{sc}
\end{table}

\begin{table}[!t]
	\caption{
    MH-pFedHN and MH-pFedHNGD experiments with different participation ratios on 50 clients, where green represents homogeneous models and red represents heterogeneous models.}
	\label{table_par}
	\begin{center}
		\begin{small}
			\begin{sc}
            \scalebox{1.0}{
				\begin{tabular}{l|cc|cc}  
					\toprule
					& \multicolumn{4}{c}{CIFAR-100}   \\ 
					\cmidrule(lr){2-5} 
					\textnormal{Methods}
					& \multicolumn{2}{c|}{\textnormal{MH-pFedHN}} 
					& \multicolumn{2}{c}{\textnormal{MH-pFedHNGD}}\\
					\cmidrule(lr){2-5} 
					Ratio \textnormal{} & \textnormal{non-IID\_1} & \textnormal{non-IID\_2}  & \textnormal{non-IID\_1} & \textnormal{non-IID\_2}  \\ 
					\midrule  
				     \rowcolor{LightGreen}
					 \textnormal{C = 20\%}  & 64.62  &77.61 & 63.89  &78.30 \\
                      \rowcolor{LightGreen}
					 \textnormal{C = 40\%}  & 64.73 & 78.49  & 65.52  & 79.15  \\ 
				
                       \rowcolor{LightGreen}
					 \textnormal{C = 60\%}  & 64.58 & 78.21& 67.02 & 78.84 \\
                     \rowcolor{LightGreen}
					 \textnormal{C = 80\%}  & 65.00 & 78.33 & 67.44 & 79.90\\
                     \rowcolor{LightGreen}
					 \textnormal{C = 100\%}  & 64.69 & 77.93 & 68.30 & 80.07\\

				    \midrule  
				    \rowcolor{LightRed}
					 \textnormal{C = 20\%} & 51.82 & 69.92 & 50.61 & 70.19  \\
                     \rowcolor{LightRed}
					 \textnormal{C = 40\%} &54.17 & 72.17 & 57.02 & 73.55  \\
				
                    \rowcolor{LightRed}
					 \textnormal{C = 60\%}  & 55.78 & 74.98 & 58.26  & 75.59  \\
                     \rowcolor{LightRed}
					 \textnormal{C = 80\%}  & 56.14  & 74.11 & 58.19 & 76.10 \\
                     \rowcolor{LightRed}
					 \textnormal{C = 100\%}  & 57.09  & 75.40 & 60.11 & 76.76 \\
				
					\bottomrule
				\end{tabular}
                }
			\end{sc}
		\end{small}
	\end{center}
\end{table}

\subsection{Experiments with Different Client Participation Ratios}
Here, we investigate the performance of MH-pFedHN and MH-pFedHNGD under partial client participation settings. The experiments are conducted on the CIFAR-100 dataset with 50 clients, where the client participation ratios are set to \{0.2, 0.4, 0.6, 0.8\}. The experimental results are shown in Table~\ref{table_par}. It can be observed that our method is robust to pFL settings with varying client participation ratios. As the participation ratio increases, the training becomes more effective, leading to better performance.

Moreover, when the participation ratio is low, we find that the performance of MH-pFedHNGD is inferior to that of MH-pFedHN in the non-IID\_1 setting. This is because non-IID\_1 is more heterogeneous than non-IID\_2. In this case, the global model can only represent the shared knowledge of a highly heterogeneous subset of clients, rather than the global shared knowledge, and thus fails to provide effective knowledge distillation guidance.

\begin{table}[!t]
	\caption{Experiment with different architectures of global model, where the model structure before $\sim$ represents the personalized client model, and the one after $\sim$ represents the global model.}
	\label{table_other_global}
	\begin{center}
		\begin{small}
			\begin{sc}
    
      \scalebox{1.0}{
        \begin{tabular}{l|ccc|ccc}  
          \toprule
          & \multicolumn{6}{c}{CIFAR-100}   \\ 
          \cmidrule(lr){2-7} 
          \textnormal{Data\_Distribution}
          & \multicolumn{3}{c|}{\textnormal{non-IID\_1}} 
          & \multicolumn{3}{c}{\textnormal{non-IID\_2}} \\
          \cmidrule(lr){2-7} 
          \# \textnormal{Clients} & 50 & 100 & 200 & 50 & 100 & 200  \\ 
          \midrule  
          \rowcolor{LightGreen}
          \textnormal{VGGNet-8$\sim$VGGNet-8}  & 63.24 & 50.67 & 44.18 & 77.88 & 77.27 & 77.54 \\
          \rowcolor{LightGreen}
          \textnormal{ResNet-10$\sim$ResNet-10}  & 61.33 & 54.66 &44.17 & 73.43 & 76.06 & 77.52 \\
          \rowcolor{LightGreen}
          \textnormal{MLP$\sim$MLP}  & 53.09 & 49.50 & 44.95 & 70.15 & 74.52 & 75.05 \\
          \rowcolor{LightRed}
          \textnormal{VGGNet-8/ResNet-10/MLP$\sim$VGGNet8}  & 59.95 & 51.01 & 44.49 & 74.39 & 77.14 & 75.97 \\
          \rowcolor{LightRed}
          \textnormal{VGGNet-8/ResNet-10/MLP$\sim$ResNet-10}  & 57.50& 51.12 & 44.59 & 74.62 & 77.02 & 76.65 \\

          \rowcolor{LightRed}
          \textnormal{VGGNet-8/ResNet-10/MLP$\sim$MLP}  & 58.33 & 49.17 & 43.94 & 74.67 & 76.58 & 76.36 \\
          \bottomrule
        \end{tabular}
      }
			\end{sc}
		\end{small}
	\end{center}
\end{table}

\begin{table}[!t]
	\caption{The experiments exploring the impact of sharing the head on the performance of MH-pFedHN and MH-pFedHNGD, where green indicates shared heads and red indicates non-shared heads.}
	\label{table_sa}
	\begin{center}
		\begin{small}
			\begin{sc}
            \scalebox{1.0}{
				\begin{tabular}{l|ccc|ccc}  
					\toprule
					& \multicolumn{6}{c}{CIFAR-100}   \\ 
					\cmidrule(lr){2-7} 
					\textnormal{Data\_Distribution}
					& \multicolumn{3}{c|}{\textnormal{non-IID\_1}} 
					& \multicolumn{3}{c}{\textnormal{non-IID\_2}}
					\\
                    \cmidrule(lr){2-7} 
					 
					\# \textnormal{Clients} & 50 & 100 & 200 & 50 & 100 & 200  \\ 
					\midrule  
                    
\rowcolor{LightGreen}
\textnormal{MH-pFedHN}  & 55.92 & 51.26 & 45.56 & 73.92 & 76.65 & 76.57 \\
\rowcolor{LightGreen}
\textnormal{MH-pFedHNGD}  & 58.22 & 53.78 & 49.83 & 75.13 & 76.89 & 77.09 \\
\rowcolor{LightRed}
\textnormal{MH-pFedHN}  & 57.84 & 51.75 & 47.03 & 74.53 & 76.73 & 76.74 \\
\rowcolor{LightRed}
\textnormal{MH-pFedHNGD}  & 58.46 & 55.43 & 50.20 & 75.23 & 77.71 & 77.28 \\

					\bottomrule
				\end{tabular}
                }
			\end{sc}
		\end{small}
	\end{center}
\end{table}

\subsection{Experiments with Different Architectures of Global Model} 

Table~\ref{table_other_global} shows the results of experiments with different architectures of the global model, which are similar to those in Table~\ref{table_ho_diff_global} when using a simple LeNet as the global model. This demonstrates the robustness of our method, as well as the effectiveness of our lightweight LeNet-based global model.

\subsection{Experiments with Shared Heads for Heterogeneous Models with Similar Parameter Sizes}

To investigate the impact of sharing the head among clients with the same number of model parameters but different structures on the performance of MH-pFedHN and MH-pFedHNGD, we adjusted the VGGNet and MLP models to match the number of parameters in the LeNet model. Consequently, the server created the same number of embedding vectors for these clients and treated their models as homogeneous, enabling all clients to share the same head. The experimental results are shown in the green section of Table~\ref{table_sa}. Additionally, we processed the clients using these three models in a manner where the number of embedding vectors varied, meaning that only clients using the same model could share the same head. The results of this setup are presented in the red section.

According to the experimental results, using different heads for models with the same number of parameters but different architectures performs slightly better than using the same head for different architectures. However, since the server generally cannot access the specific model architecture information of the clients, our method is sufficiently robust; under the premise of protecting model structure privacy, it achieves a reasonable balance between personalized performance and privacy protection.

\begin{table}[!t]
    \centering
    \caption{Communication and computation overhead under non-IID\_1, the results are similar in two non-IID settings. Green indicates MH-pFedHN, red indicates MH-pFedHNGD.}
    \label{table:comm_and_comp_overhead}
    \begin{tabular}{lcc}
        \toprule
        Algorithm & Overall Computation (mins) & Overall Communication (MB) \\
        \midrule
         \rowcolor{LightGreen}
         homogeneous   &  77.01  &  1.83 \\
          \rowcolor{LightGreen}
          heterogeneous &   303.8 &  3.00 \\
           \midrule  
           \rowcolor{LightRed}
           homogeneous   &  133.7  &  3.66 \\
            \rowcolor{LightRed}
          heterogeneous &  407.8  & 4.83 \\
        \bottomrule
    \end{tabular}
\end{table}

\subsection{Experiments with Communication and Computation Overhead}

Table~\ref{table:comm_and_comp_overhead} shows the communication and computation overhead of MH-pFedHN and MH-pFedHNGD for 50 clients with 500 rounds on CIFAR-100. The computation overhead of MH-pFedHNGD is 2.0 and 1.61 times that of MH-pFedHN in homogeneous and heterogeneous settings, respectively, which is consistent with our experimental setup. The computation overhead for MH-pFedHNGD is higher than that of MH-pFedHN by 73\% and 34\% in homogeneous and heterogeneous settings, respectively. This indicates that our lightweight global model introduces only minimal additional time complexity while significantly enhancing the generalization ability of the hypernetwork and improving overall performance, especially for heterogeneous settings.

\begin{table}[!t]
    \caption{We used CIFAR-100 to evaluate the impact on MH-pFedHN and MH-pFedHNGD of only accepting the first 30\% weight update with the largest absolute value, where green and red correspond to the homogeneous and the heterogeneous models, respectively.}
    \label{table_imp}
    \begin{center}
        \begin{small}
            \begin{sc}
                \scalebox{1.0}{
                    \begin{tabular}{l|ccc|ccc}  
                        \toprule
                        & \multicolumn{6}{c}{CIFAR-100}   \\ 
                        \cmidrule(lr){2-7} 
                        \textnormal{Data\_Distribution}
                        & \multicolumn{3}{c|}{\textnormal{non-IID\_1}} 
                        & \multicolumn{3}{c}{\textnormal{non-IID\_2}}\\
                        \cmidrule(lr){2-7} 
                        \# \textnormal{Clients} & 50 & 100 & 200 & 50 & 100 & 200  \\ 
                        \midrule  
                        \rowcolor{LightGreen}
                        \textnormal{MH-pFedHN}  & 63.91  &63.14  & 59.25  &78.25  & 80.79 & 81.04 \\
                        \rowcolor{LightGreen}
                        \textnormal{Top 30\%}  & 64.35 & 62.97  & 58.39  & 78.78  & 80.91& 81.10 \\ 
                        \cmidrule(lr){1-7} 
                        \rowcolor{LightGreen}
                        \textnormal{MH-pFedHNGD}  & 68.27 & 63.58 & 61.19& 80.18 & 82.31 & 82.19 \\
                        \rowcolor{LightGreen}
                        \textnormal{Top 30\%}  & 67.79 & 63.96 & 61.56 & 80.22 &83.00 & 82.92\\
                        \midrule  
                        \rowcolor{LightRed}
                        \textnormal{MH-pFedHN} & 57.18 & 50.35 & 40.83 & 74.84  & 76.78 & 74.71  \\
                        \rowcolor{LightRed}
                        \textnormal{Top 30\%} &57.35 & 47.61 & 40.68 & 75.33  & 77.11 & 75.35 \\
                        \cmidrule(lr){1-7} 
                        \rowcolor{LightRed}
                        \textnormal{MH-pFedHNGD}  & 60.11 & 51.57  & 43.41 & 76.76  & 77.03 & 76.46 \\
                        \rowcolor{LightRed}
                        \textnormal{Top 30\%}  & 59.39  & 50.80 & 43.55 & 75.97 & 77.55 & 75.68 \\
                        \bottomrule
                    \end{tabular}
                }
            \end{sc}
        \end{small}
    \end{center}
\end{table}

\subsection{Experiments with Communication Efficiency using Weight Pruning}
Due to the introduction of a global model in MH-pFedHNGD, although its number of parameters is the same as the minimum number of parameters required by the client, it still increases communication overhead. We considered weight pruning, which involves pruning the weight updates of personalized models uploaded by the client to the server and only uploading the top 30\% of updates with the largest absolute values in each round of communication. We also applied this pruning method to MH-pFedHN, with the experimental results presented in Table~\ref{table_imp}. Additionally, specific parameter details can be found in Appendix~\ref{append:ModelArchitectures}.

Unexpectedly, weight pruning did not significantly decrease accuracy. The model's performance is comparable to that when no pruning is applied. This indicates that the hypernetwork of our methods can still learn efficiently and maintain high performance even with less information, and has great potential for practical use, effectively retaining key information and ensuring that model accuracy is not compromised.

\begin{figure}[!t]
    \centering
    \includegraphics[width=0.6\textwidth]{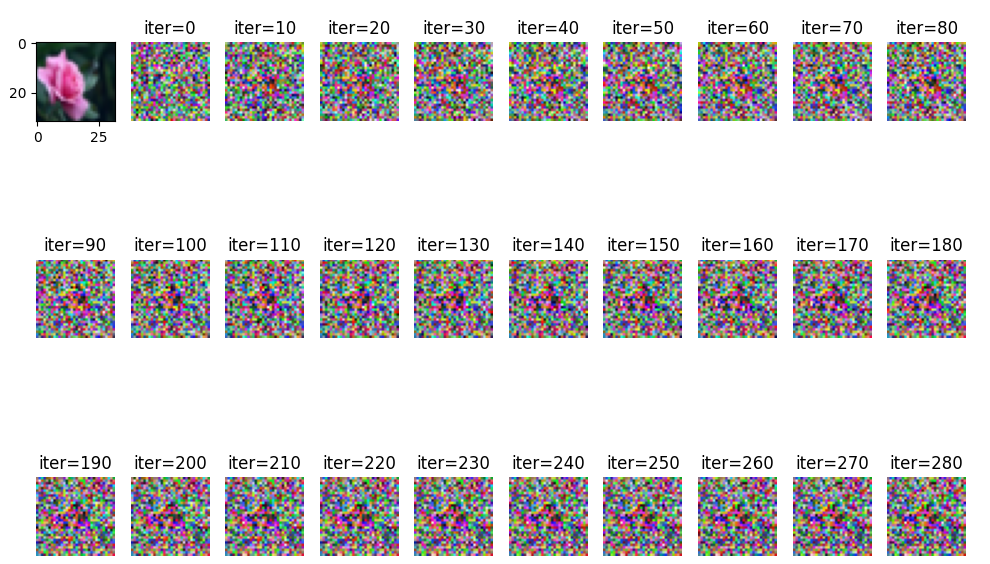} 
    \caption{iDLG Gradient Inversion Attack  using MH-pFedHNGD}
    \label{fig_attack}
\end{figure}

\subsection{Experiments with iDLG Attacks using MH-pFedHNGD}
Figure~\ref{fig_attack} shows the results of iDLG~\cite{Zhao2020iDLGID} using MH-pFedHNGD, which exhibits that our methods could still preserve the data security for clients even using the plug-in component global model.

\begin{table}[!t]
\caption{Experiment with extreme personalized setting on CIFAR-100.}
\label{table_gen_p}
\begin{center}
\begin{small}
\begin{sc}
\begin{tabular}{lccccccr}  
\toprule
&& \multicolumn{6}{c}{CIFAR-100}   \\ 
\cmidrule(lr){2-8} 
&\textnormal{Data\_Distribution}
& \multicolumn{3}{c}{\textnormal{non-IID\_1}} 
& \multicolumn{3}{c}{\textnormal{non-IID\_2}}\\
\cmidrule(lr){3-5} 
\cmidrule(lr){6-8} 
&\# \textnormal{Clients} & 50 & 100 & 200 & 50 & 100 & 200  \\ 
\midrule  
\multirow{2}{*}{\textnormal{\makecell{Homogeneous model}}}  				
&\textnormal{MH-pFedHN}  & 56.96  &51.62  & 44.38 & 75.59  & 77.71  & 78.53 \\
&\textnormal{MH-pFedHNGD}  & 59.67 & 56.20 & 49.23 & 76.58  &79.09 & 79.99  \\
\midrule 
\multirow{2}{*}{\textnormal{\makecell{Heterogeneous model}}}  	
&\textnormal{MH-pFedHN}  & 54.37  & 46.00& 38.80 & 74.49  & 76.32 & 75.38 \\
&\textnormal{MH-pFedHNGD}  & 58.71 & 50.07  & 41.79  & 77.23& 78.16  & 76.59  \\
\bottomrule
\end{tabular}
\end{sc}
\end{small}
\end{center}
\end{table}

\subsection{Experiments with Extreme Personalized Setting}

Here, we fully consider the generation of personalized parameters for clients, which allows clients to choose to retain parameters for certain layers locally without participating in federated training under the extremely personalized setting. This greatly satisfies the personalized needs of the clients. 

Following the settings in Section~\ref{Homogeneous Model} and~\ref{Heterogeneous Model}, in the experiments with homogeneous models, all clients use a LeNet-style model. We have set up four modes, 1/4 of the clients are required to generate parameters for the input layer and hidden layers, while retaining the parameters for the classification layer locally; 1/4 of the clients are required to generate parameters for the hidden layers, retaining the parameters for the input and classification layers locally; 1/4 of the clients are required to generate parameters for the hidden and classification layers, retaining the parameters for the input layer locally; and the final 1/4 of the clients are required to generate parameters for all layers.   
In the experiments with heterogeneous models, the setup for the VGG model is similar to that of the LeNet-style model. For the residual networks, we configure some clients to generate only the parameters for the convolutional layers, retaining the parameters for the batch normalization layers and classification layers locally, while other clients are required to generate all parameters. Therefore, in the experiments with heterogeneous models, there are 14 different modes, with the number of clients in each mode being 1/14 of the total. 

The experimental results, as shown in Table~\ref{table_gen_p}, indicate that when client personalization needs are met, the accuracy decreases to varying degrees. Therefore, a trade-off between personalization and performance is necessary.

\begin{table}[!t]
\caption{MH-pFedHN in resource-constrained experiments. Test accuracy over 50, 100, and 200 clients on the CIFAR-100.}
\label{table_gen_c}
\begin{center}
\begin{small}
\begin{sc}
\begin{tabular}{lccccccr}  
\toprule
&& \multicolumn{6}{c}{CIFAR-100}   \\ 
\cmidrule(lr){2-8} 
&\textnormal{Data\_Distribution}
& \multicolumn{3}{c}{\textnormal{non-IID\_1}} 
& \multicolumn{3}{c}{\textnormal{non-IID\_2}}\\
\cmidrule(lr){3-5} 
\cmidrule(lr){6-8} 
&\# \textnormal{Clients} & 50 & 100 & 200 & 50 & 100 & 200  \\ 
\midrule  
\multirow{1}{*}{\textnormal{\makecell{Homogeneous model}}}  				
&\textnormal{MH-pFedHN}  & 53.61  &45.30  & 40.93 & 73.91 & 77.37  & 77.78\\
\midrule 
\multirow{1}{*}{\textnormal{\makecell{Heterogeneous model}}}  	
&\textnormal{MH-pFedHN}  & 53.28  & 46.08  & 37.67  & 75.10  & 76.72 & 76.01 \\
\bottomrule
\end{tabular}
\end{sc}
\end{small}	
\end{center}
\end{table}

\begin{table}[!t]
	\caption{MH-pFedHNGD in resource-constrained experiments, where green represents homogeneous models and red represents heterogeneous models.}
	\label{table_re_MH-pFedHNGD}
	
	\begin{center}
		\begin{small}
			\begin{sc}
            \scalebox{1.0}{
				\begin{tabular}{l|ccc|ccc}  
					\toprule
					& \multicolumn{6}{c}{CIFAR-100}   \\ 
					\cmidrule(lr){2-7} 
					\textnormal{Methods}
					& \multicolumn{3}{c|}{\textnormal{non-IID\_1}} 
					& \multicolumn{3}{c}{\textnormal{non-IID\_2}}\\
					\cmidrule(lr){2-7} 
					\# \textnormal{Clients} & \textnormal{50} & \textnormal{100}  & \textnormal{200} & \textnormal{50}& \textnormal{100}& \textnormal{200}  \\ 
					\midrule  
                    \rowcolor{LightGreen}
					 \textnormal{C = 0\%}  & 64.69  &63.32 & 60.11  &77.93 & 80.93  &81.60 \\
				     \rowcolor{LightGreen}
					 \textnormal{C = 20\%}  & 64.82  &63.41 & 60.42  &78.87 & 80.91  &81.93 \\
                      \rowcolor{LightGreen}
					 \textnormal{C = 40\%}   & 65.17  &63.07 & 60.34  &78.63 & 81.85  &82.48  \\ 
				
                       \rowcolor{LightGreen}
					 \textnormal{C = 60\%}  & 66.48  &62.60 & 61.05  &79.22 & 82.05  &82.59 \\
                     \rowcolor{LightGreen}
					 \textnormal{C = 80\%}   & 66.50  &63.38 & 61.33  &79.43 & 82.77  &82.79\\

                     \rowcolor{LightGreen}
					 \textnormal{C = 100\%}   & 68.30  &63.97 & 61.59  &80.07 & 82.51  &82.54\\
                     
				    \midrule  
                     \rowcolor{LightRed}
					 \textnormal{C = 0\%} & 57.09  &50.53 & 42.98  &75.40 &77.24  &75.38  \\
				    \rowcolor{LightRed}
					 \textnormal{C = 20\%} & 57.48  &50.23 & 42.55  &75.90 &76.95  &75.21  \\
                     \rowcolor{LightRed}
					 \textnormal{C = 40\%}  & 58.48  &49.17 & 41.11  &74.90 & 77.03  &75.07\\
				
                    \rowcolor{LightRed}
					 \textnormal{C = 60\%}  & 58.06  &51.42 & 42.64  &75.56 & 76.79  &74.90  \\
                     \rowcolor{LightRed}
					 \textnormal{C = 80\%}   & 58.79  &50.63 & 37.38  &76.44 & 77.51  &76.22\\
                     \rowcolor{LightRed}
					 \textnormal{C = 100\%}   & 60.11  &52.14 & 43.41  &76.76 & 77.03  &76.46\\
				
					\bottomrule
				\end{tabular}
                }
			\end{sc}
		\end{small}
	\end{center}

\end{table}

\subsection{Experiments under Resource Constraint Setting}

Here, we first investigate the performance of MH-pFedHN under resource-constrained conditions. In resource-constrained environments, such as on-edge devices or mobile platforms, clients often face limitations in computational power, memory, and storage capacity, significantly impacting their ability to effectively train deep learning models. Training a complete model with all layers requires substantial computational resources, which may be beyond the capabilities of these clients. Additionally, deep neural networks consume significant amounts of memory, making it challenging to load an entire model, especially for clients with limited memory capacity. Given these constraints, clients can only feasibly train a subset of the model, typically focusing on the shallower layers. This approach reduces the number of parameters, thereby lowering the memory and computational requirements.

To simulate such a scenario, we adhered to the configurations outlined in Sections~\ref{Homogeneous Model} and~\ref{Heterogeneous Model} of our study. In the experiments with homogeneous models, all clients use a LeNet-style model. We have set up three modes. In the first setting, we froze the parameters of the second fully connected layer and the classification layer, simulating the most severe resource constraints. In the second setting, we froze only the parameters of the classification layer. In the third setting, we allowed clients without resource limitations to train the entire model. These three settings were evenly distributed among the clients. In experiments with heterogeneous models, the VGG model's configuration was similar to that of the LeNet-style model. For residual networks, we established two scenarios: in the first scenario, we froze the parameters of the last 20\% of the layers to mimic resource-constrained situations. In the second scenario, clients had no resource limitations and could therefore train the entire model. Likewise, these 12 different configurations were also evenly distributed among the clients. This design allows us to assess the performance variations in different resource availability situations, providing insights into how model architecture and training strategies influence overall effectiveness in heterogeneous client environments. 

The experimental results are summarized in Table~\ref{table_gen_c}. We can observe that under resource-constrained conditions, the accuracy decreases to varying degrees across the different configurations. In particular, in the case of the non-IID\_1 distribution, the decrease in accuracy is more pronounced. This can be attributed to the greater heterogeneity of the non-IID\_1 scenario, where, on average, each client receives a smaller amount of data.

Next, we conducted an in-depth study of the performance of the MH-pFedHNGD algorithm in resource-constrained environments. In such scenarios, only a portion of clients can deploy the global model. We set the deployment ratios to 20\%, 40\%, 60\%, and 80\% of clients, respectively, to systematically evaluate the impact of different deployment scales on overall performance. As shown in Table~\ref{table_re_MH-pFedHNGD}, the analysis of the experimental data reveals that when the proportion of clients deploying the global model reaches 40\%, there is a significant improvement in system performance. This indicates that in resource-limited environments, not all clients need to deploy the global model; as long as a certain proportion of clients can utilize the global model, the overall performance of the federated learning system can be effectively enhanced. This finding provides important guidance for optimizing resource allocation and improving system efficiency in practical applications.

\section{Experiment with Hyperparameter Choices}
\label{append:designlabels} 

\begin{figure}[!t]
        \begin{center}
            \centerline{\includegraphics[width=0.5\linewidth]{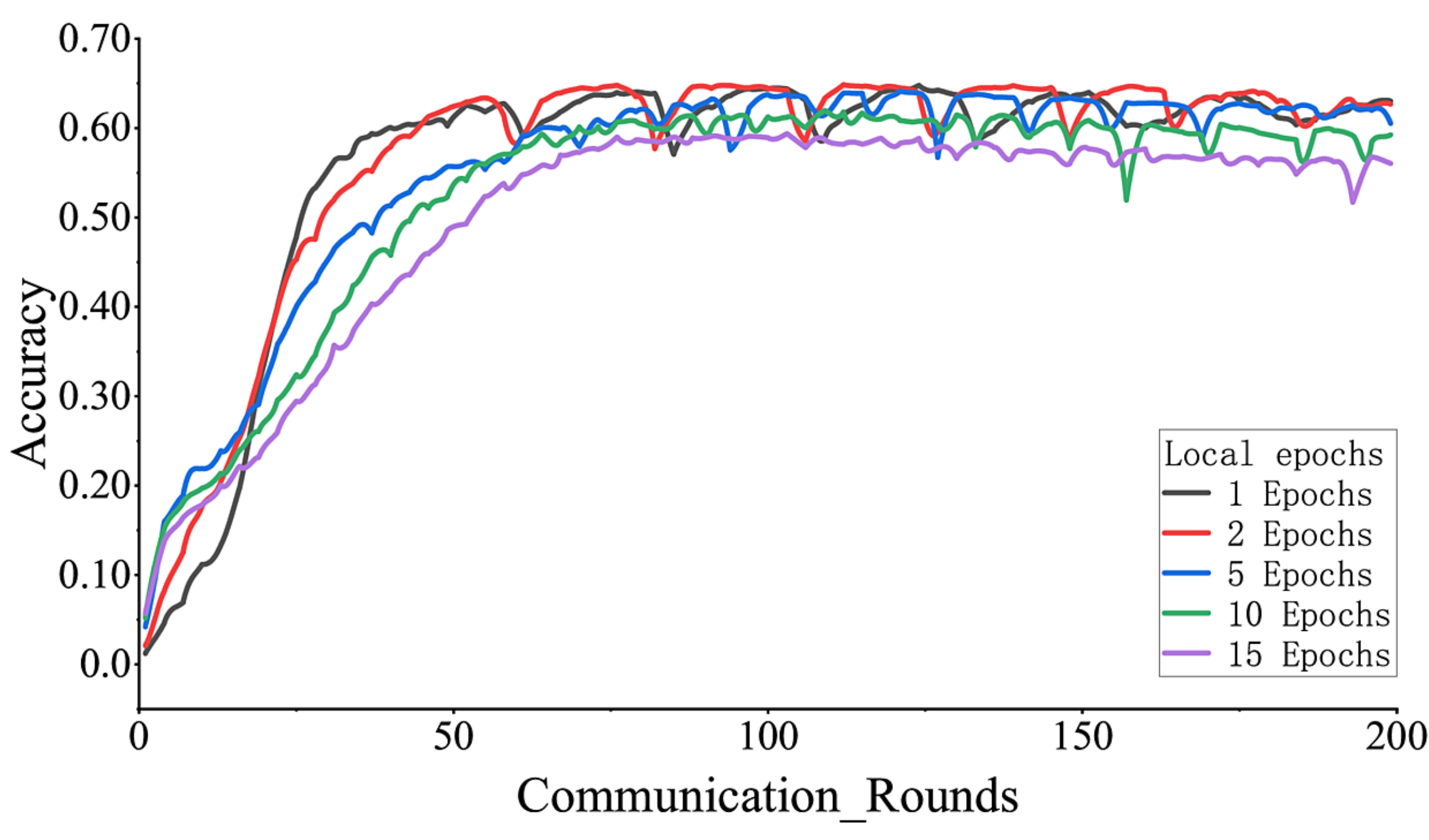}}
            \caption{Experiment with the number of local epochs on the CIFAR-100 dataset using MH-pFedHN.}
        \label{epoch}
        \end{center}
\end{figure}

\begin{figure}[!t]  
	\centering  
	\subfigure[]{  
		\includegraphics[scale=0.06]{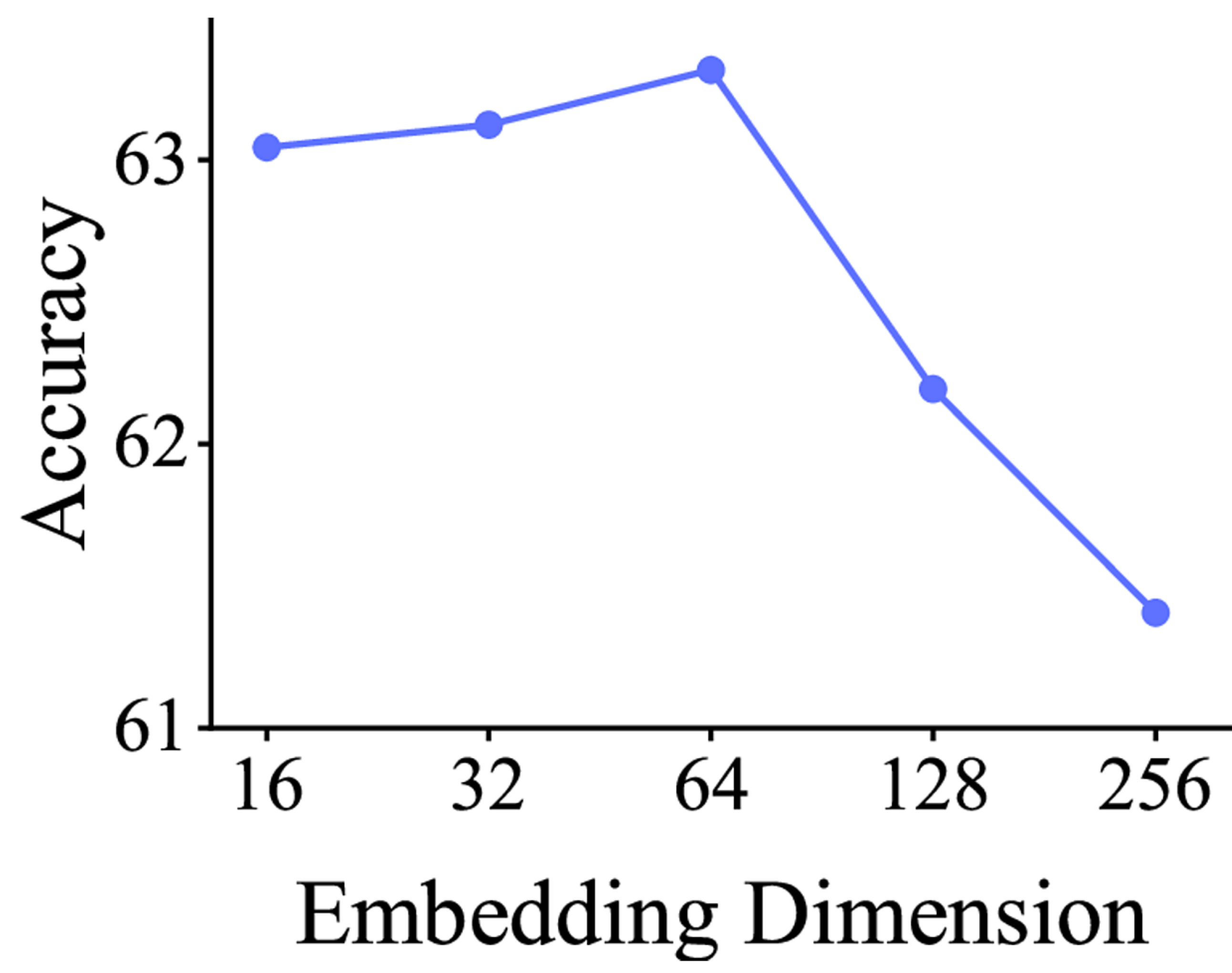}  
		\label{emb_dim}  
	}  
	\hspace{-5mm}
	\subfigure[]{  
		\includegraphics[scale=0.065]{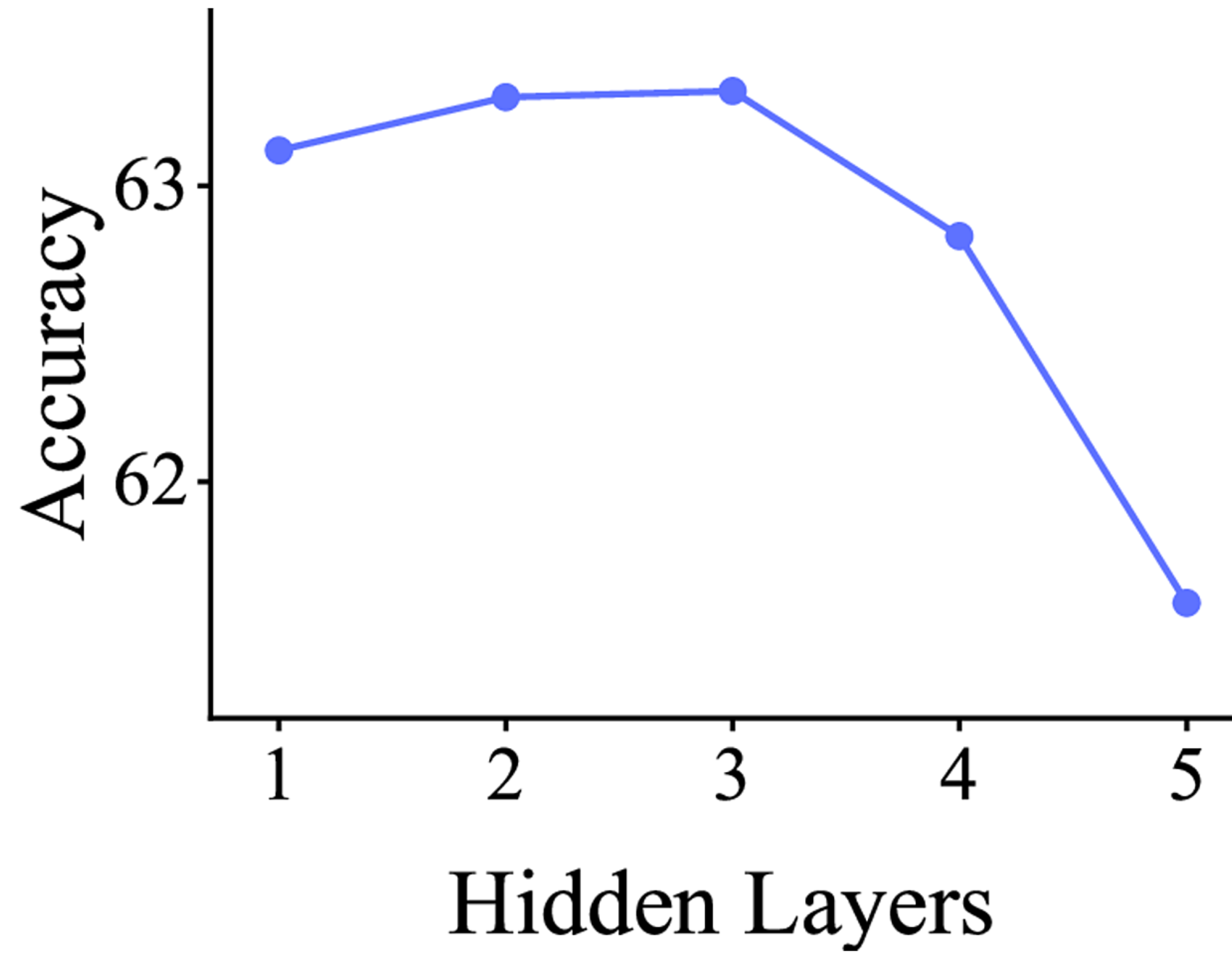}  
		\label{cnn_hnet_dim}  
	}  
	\hspace{-5mm}
	\subfigure[]{  
		\includegraphics[scale=0.065]{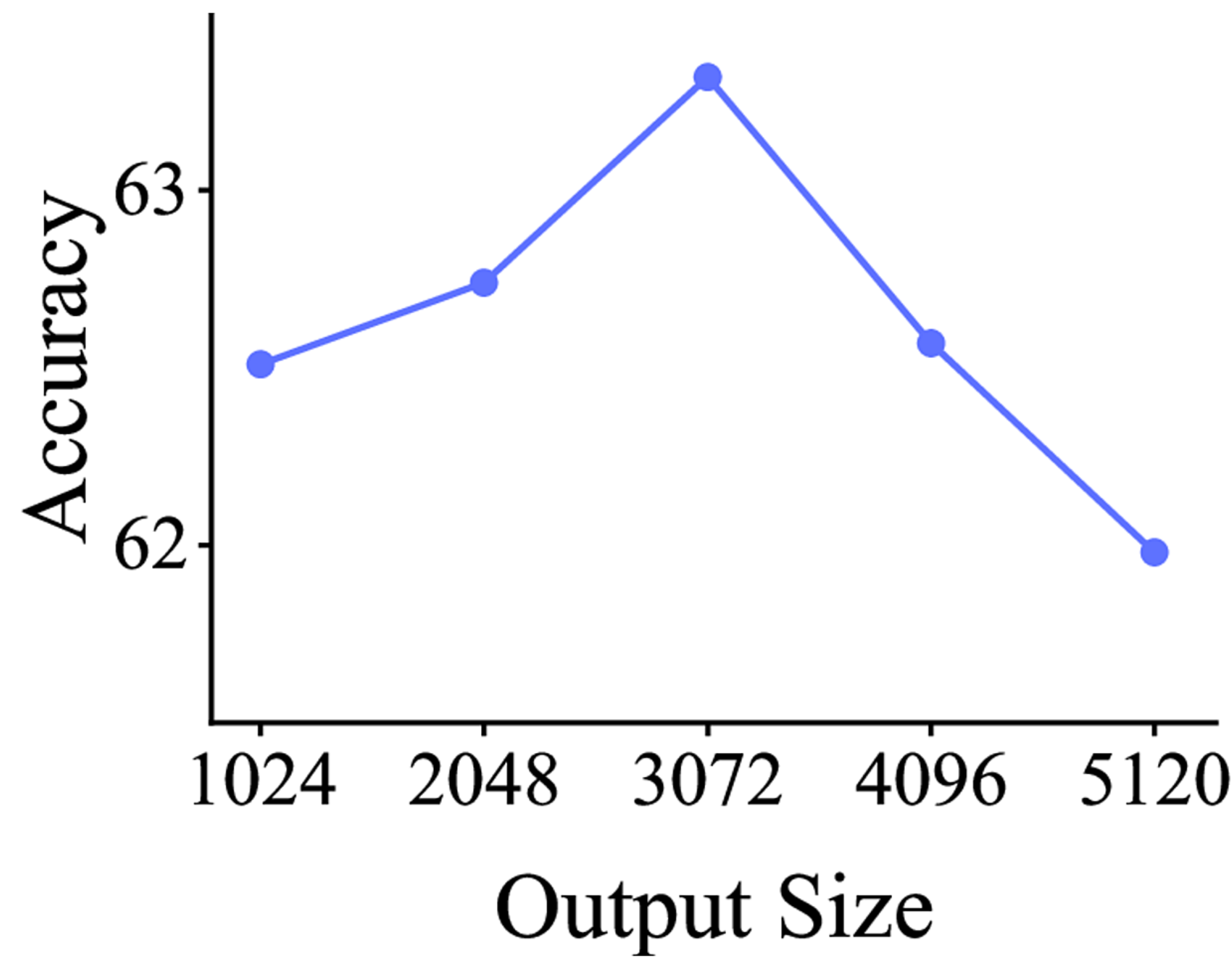}  
		\label{hnet_size}  
	}  
        \hspace{-5mm}
        \subfigure[]{  
		\includegraphics[scale=0.065]{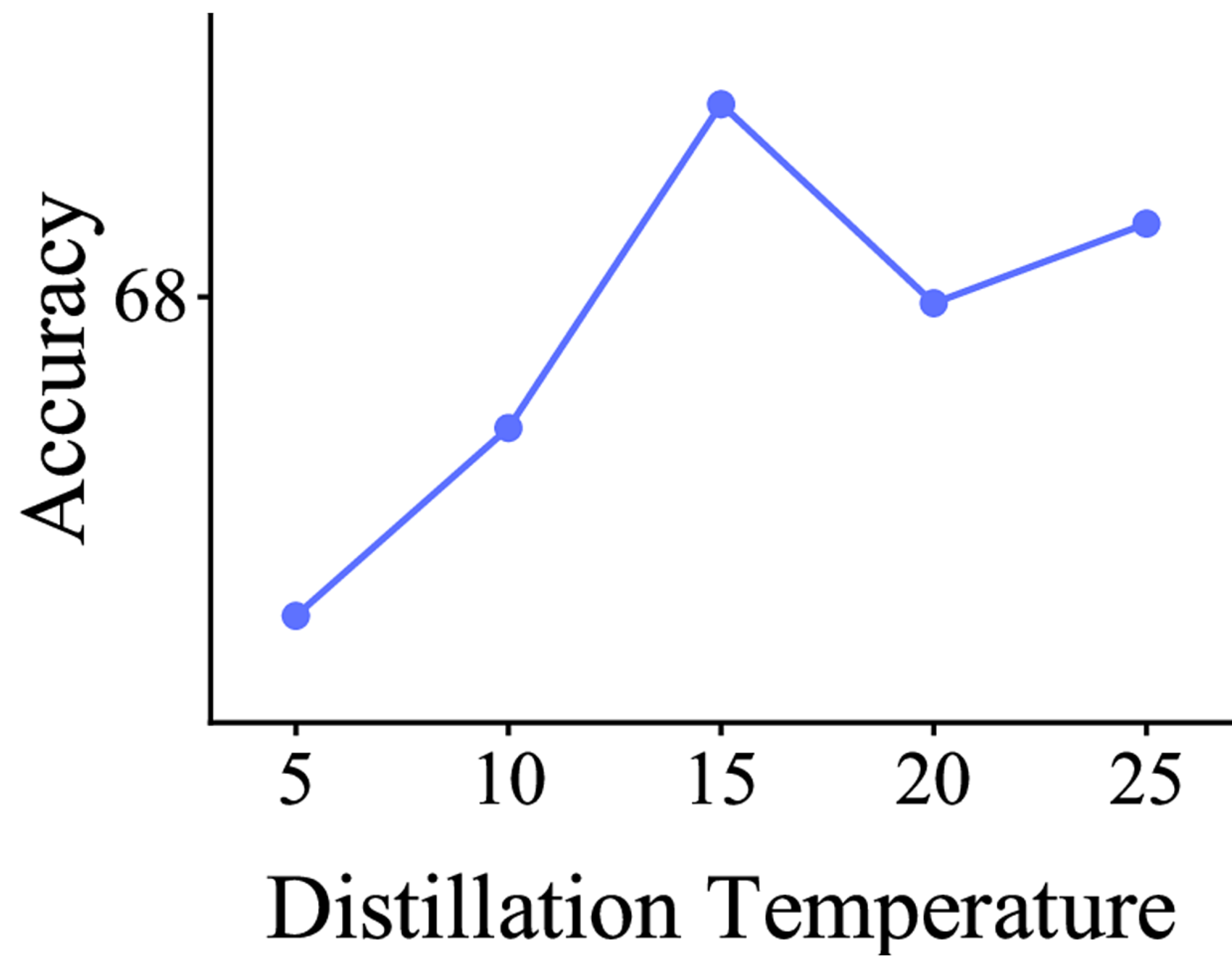}  
		\label{cnn_dt}  
	}  
        \hspace{-5mm}
        \subfigure[]{  
		\includegraphics[scale=0.065]{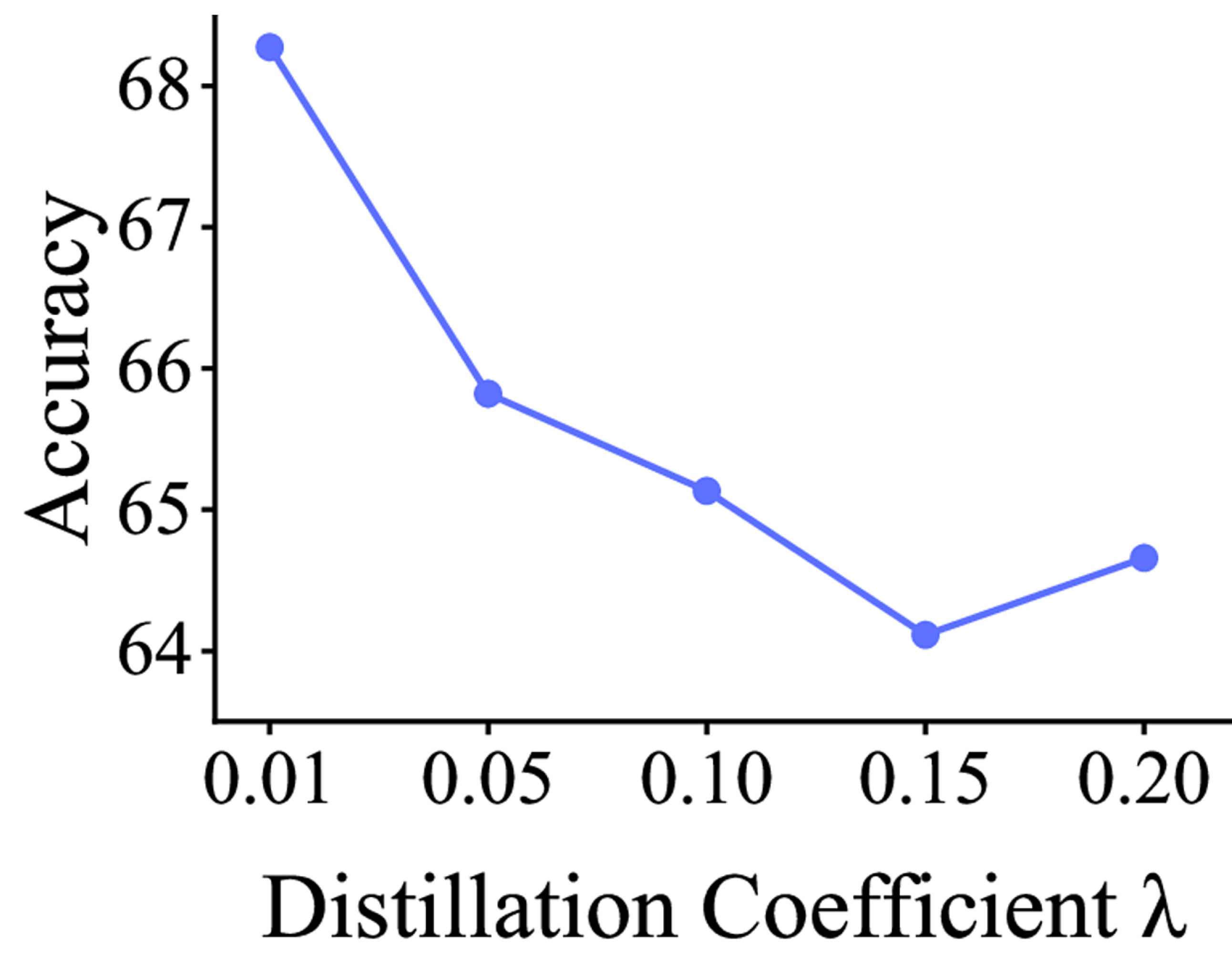}  
		\label{cnn_da}  
	} 
	\caption{MH-pFedHN test results on CIFAR-100 showing the effect of (a) the dimension of client embedding vector, (b) the number of hypernetwork's hidden layers, and (c) the output size of hypernetworks, (d) the distillation temperature with MH-pFedHNGD and (e) the balancing factor on distillation loss with MH-pFedHNGD.}  
	\label{fig}  
\end{figure}

\subsection{Experiments with Number of Local Epochs}
Here, we set local epochs to \{1,2,5,10,15\}. Figure~\ref{epoch} shows the test accuracy throughout training over 200 rounds. Results indicate that MH-pFedHN is relatively robust to the choice of local optimization steps. 


\subsection{Experiments with Client Embedding Dimension}
We investigate the effect of the dimension of the embedding vector on the performance of MH-pFedHN. Specifically, we run an ablation study on a set of different embedding dimensions \{16, 32, 64, 128, 256\}. The results are presented in Figure~\ref{emb_dim}. We show MH-pFedHN robustness to the dimension of the client embedding vector; hence, we fix the embedding dimension through all experiments to 64.

\subsection{Experiments with Hypernetwork Capacity}
Here, we inspect the effect of the Hypernetwork's capacity on the local network performance. We conducted an experiment in which we changed the depth of the Hypernetwork by stacking fully connected layers.
We evaluate MH-pFedHN using different hidden layers \{1, 2, 3, 4, 5\}. Figure~\ref{cnn_hnet_dim} presents the final test accuracy. MH-pFedHN achieves optimal performance with three hidden layers. We use three hidden layers of HN for all experiments in the main text.

\subsection{Experiments with Output Size of Hypernetworks}
We investigate how the output size of hypernetworks impacts the performance of local networks. In our experimentation, we adjusted the output size to see its effects. We evaluated MH-pFedHN in various output sizes \{1024, 2048, 3072, 4096, 5120\}. The findings presented in Figure~\ref{hnet_size} indicate that MH-pFedHN performs best with an output size of 3072, which we use for all subsequent experiments in the main text.

\subsection{Experiments with Temperature for Knowledge Distillation}
We examine the influence of the temperature parameter, which is crucial in knowledge distillation, on the performance of the MH-pFedHNGD framework using the CIFAR-100 dataset. We systematically varied the temperature settings to investigate their impact on the efficiency of MH-pFedHNGD. The specific temperature values tested are \{5, 10, 15, 20, 25\}. The results, illustrated in Figure~\ref{cnn_dt}, demonstrate that MH-pFedHNGD achieves optimal performance with a temperature of 15 in the CIFAR-100 dataset under the non-IID\_1 setting. This optimal temperature enhances the softmax distribution, thereby improving the transfer of knowledge from the global model to the private model within MH-pFedHNGD. Therefore, we utilize this temperature setting for all experiments conducted on the CIFAR-100 dataset.

\subsection{Experiments with Balancing Factor on Distillation Loss}
We examine the influence of the balancing factor in the loss function, which is critical in knowledge distillation, on the performance of the MH-pFedHNGD framework using the CIFAR-100 dataset. We systematically varied the ratio between the true loss and the distillation loss to investigate their impact on the efficiency of MH-pFedHNGD. The specific balancing factor values tested were \{0.01, 0.05, 0.1, 0.15, 0.2\}.
The results, illustrated in Figure~\ref{cnn_da}, demonstrate that MH-pFedHNGD achieves optimal performance with a balancing factor of 0.01 in the CIFAR-100 dataset under the non-IID\_1 setting. This optimal ratio facilitates an effective trade-off between accurately modeling the true data distribution and leveraging the global model's knowledge, thereby improving the overall learning process. Consequently, we utilize this balancing factor setting for all experiments conducted on the CIFAR-100 dataset.

\begin{table}[!t]  
	\centering  
	\caption{LeNet-style Model Structure}  
	\renewcommand{\arraystretch}{1.5} 
     \scalebox{0.8}{
	\begin{tabular}{@{}>{\centering\arraybackslash}p{2.5cm}   
			>{\centering\arraybackslash}p{3cm}   
			>{\centering\arraybackslash}p{3cm}@{}}  
		\toprule  
		\textbf{Layer} & \textbf{Shape} & \textbf{Nonlinearity} \\   
		\midrule  
		Conv1 & $3 \times 3 \times 3 \times 16$ & ReLU \\   
		MaxPool & $2 \times 2$ & - \\   
		Conv2 & $16 \times 3 \times 3 \times 32$ & ReLU \\   
		MaxPool & $2 \times 2$ & Flatten \\   
		FC1 & $2048 \times 108$ & ReLU \\   
		FC2 & $108 \times 64$ & ReLU \\   
		FC3 & $64 \times 100$ & None \\   
		\bottomrule  
	\end{tabular}  
    }
	\label{CNN}  
\end{table}  
\begin{table}[!t]  
	\centering  
	\caption{MLP Model Structure}  
	\renewcommand{\arraystretch}{1.5} 
    \scalebox{0.8}{
	\begin{tabular}{@{}>{\centering\arraybackslash}p{3cm}   
			>{\centering\arraybackslash}p{3cm}   
			>{\centering\arraybackslash}p{3cm}@{}}  
		\toprule  
		\textbf{Layer} & \textbf{Shape} & \textbf{Nonlinearity} \\   
		\midrule   
		FC1 & $3072 \times 128$ & ReLU \\   
		FC2 & $128 \times 64$ & ReLU \\   
		FC3 & $64 \times 100$ & None \\   
		\bottomrule  
	\end{tabular}  
    }
	\label{MLP}  
\end{table}  
\begin{table}[!t]  
	\centering  
	\caption{Simplified VGG8 Model Structure}  
	\renewcommand{\arraystretch}{1.5} 
    \scalebox{0.8}{
	\begin{tabular}{@{}>{\centering\arraybackslash}p{3cm}   
			>{\centering\arraybackslash}p{3cm}   
			>{\centering\arraybackslash}p{3cm}@{}}  
		\toprule  
		\textbf{Layer} & \textbf{Shape} & \textbf{Nonlinearity} \\   
		\midrule  
		Conv1 & $3 \times 3 \times 3 \times 16$ & ReLU \\   
		Conv2 & $16 \times 3 \times 3 \times 16$ & ReLU \\   
		MaxPool & $2 \times 2$ & - \\   
		Conv3 & $16 \times 3 \times 3 \times 32$ & ReLU \\   
		Conv4 & $32 \times 3 \times 3 \times 32$ & ReLU \\   
		MaxPool & $2 \times 2$ & - \\   
		Conv5 & $32 \times 3 \times 3 \times 64$ & ReLU \\   
		Conv6 & $64 \times 3 \times 3 \times 64$ & ReLU \\   
		MaxPool & $2 \times 2$ & Flatten \\   
		Linear1 & $1024 \times 180$ & ReLU \\   
		Linear2 & $108 \times 64$ & ReLU \\   
		Linear3 & $64 \times 100$ & None \\   
		\bottomrule  
	\end{tabular}  
    }
	\label{VGG8}  
\end{table}  
\begin{table}[!t]  
	\centering 
	\caption{Structure of three Residual Networks Models}  
	\renewcommand{\arraystretch}{1.5} 
        \scalebox{0.75}{
	\begin{tabular}{@{}>{\centering\arraybackslash}p{2.5cm}   
			>{\centering\arraybackslash}p{2cm}   
			>{\centering\arraybackslash}p{4cm}@{}
			>{\centering\arraybackslash}p{4cm}@{}
			>{\centering\arraybackslash}p{4cm}@{}}  
		\toprule  
		\textbf{Group Name} & \textbf{Output Size} & \textbf{10-layer ResNet} & \textbf{12-layer ResNet} & \textbf{18-layer ResNet}\\   
		\midrule  
		Conv1 & $32 \times 32$ & $[3 \times 3, 16]$ & $[3 \times 3, 16]$ & $[3 \times 3, 16]$ \\   
		Conv2 & $32 \times 32$ & $\left[\begin{array}{c} 3 \times 3, 16 \\[0.5ex] 3 \times 3, 16 \end{array}\right]\times 3 $ & $\left[\begin{array}{c} 3 \times 3, 16 \\[0.5ex] 3 \times 3, 16 \end{array}\right]\times 1$ & $\left[\begin{array}{c} 3 \times 3, 16 \\[0.5ex] 3 \times 3, 16 \end{array}\right]\times 6 $ \\   
		Conv3 & $16 \times 16$ & $\left[\begin{array}{c} 3 \times 3, 32 \\[0.5ex] 3 \times 3, 32 \end{array}\right]\times 3$ & $\left[\begin{array}{c} 3 \times 3, 32 \\[0.5ex] 3 \times 3, 32 \end{array}\right]\times 5$ & $\left[\begin{array}{c} 3 \times 3, 32 \\[0.5ex] 3 \times 3, 32 \end{array}\right]\times 6$ \\   
		Conv4 & $8 \times 8$ & $\left[\begin{array}{c} 3 \times 3, 64 \\[0.5ex] 3 \times 3, 64 \end{array}\right]\times 4$ & $\left[\begin{array}{c} 3 \times 3, 64 \\[0.5ex] 3 \times 3, 64 \end{array}\right]\times 6$ & $\left[\begin{array}{c} 3 \times 3, 64 \\[0.5ex] 3 \times 3, 64 \end{array}\right]\times 6$ \\ 
		Avg-Pool & $1 \times 1$ & $[8 \times 8]$ & $[8 \times 8]$ & $[8 \times 8]$ \\   
		\bottomrule  
	\end{tabular}  
        }
	\label{ResNet}  
\end{table}  
\begin{table}[!t]
    \centering
    \caption{Comparison of Model Parameter Sizes Uploaded Before and After Weight Pruning Across Different Models in the CIFAR-100 Task.}
    \scalebox{1.1}{
    \begin{tabular}{@{}ccc@{}}

        \toprule
        Model      & Params   & Top 30\%    \\ \midrule
        LeNet-style Model        & 0.915 M & 0.274M  \\
        Simplified VGG8       & 1.048 M & 0.314M  \\
        10-layer ResNet   & 1.347 M & 0.404M  \\
        12-layer ResNet   & 2.018 M & 0.605M  \\
        18-layer ResNet   & 2.179 M & 0.654M  \\ \bottomrule
    \end{tabular}
    }
    \label{tab:model_params}
\end{table}

\section{Model Architectures and Parameters}
\label{append:ModelArchitectures} 
Here, we present all the model architectures and the parameters used in the CIFAR-100 experiments. Table~\ref{CNN} is a LeNet-style model, Table~\ref{MLP} is an MLP model, Table~\ref{VGG8} is a simplified VGG model (8 layers), Table~\ref{ResNet} is three residual networks, Table~\ref{tab:model_params} lists the parameters of all models as well as those that need to be uploaded after pruning.

\end{document}